%% file: iclr2026_conference.tex
\documentclass{article} %
\usepackage{iclr2026_conference,times}

\input{math_commands.tex}

\usepackage{hyperref}
\usepackage{url}

\usepackage[utf8]{inputenc} %
\usepackage[T1]{fontenc}    %
\usepackage{hyperref}       %
\usepackage{url}            %
\usepackage{booktabs}       %
\usepackage{amsfonts}       %
\usepackage{nicefrac}       %
\usepackage{microtype}      %
\usepackage{xcolor}         %
\usepackage{listings}
\usepackage{multirow}
\usepackage{amsmath}
\usepackage{colortbl}
\usepackage{enumitem}
\usepackage{xspace}
\usepackage{wrapfig}
\usepackage{graphicx}
\usepackage{pifont}

\renewcommand{\cite}{\citep}

\newcommand\llama{\raisebox{-7pt}{\includegraphics[width=1em]{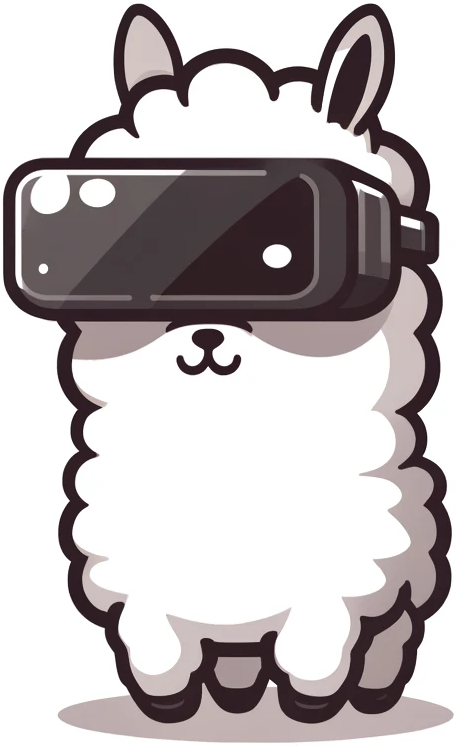}}}

\newcommand{\eg}{\textit{e.g.}}

\newcommand{\methodname}{ComboBench\xspace}
\newcommand{\numberofscenarios}{262\xspace}
\newcommand{\numberofgames}{four\xspace}
\newcommand{\numberofllms}{six\xspace}
\newcommand{\numberofhiredplayers}{eight\xspace}

\definecolor{mygray}{RGB}{226, 226, 226}
\definecolor{myred}{RGB}{252, 142, 142}
\definecolor{mygreen}{RGB}{147, 255, 143}
\definecolor{myblue}{RGB}{144, 155, 255}
\definecolor{myyellow}{RGB}{253, 253, 143}
\definecolor{mypurple}{RGB}{255, 142, 250}

\lstdefinelanguage{dsl}{
    morekeywords={program, statement_list, statement, time_constrained_statement, control_flow_statement, manipulation_statement, variable_declaration_statement, device, property, value, pose_property, orientation_property, button_property, axis_property, position_coordinate, orientation_coordinate, referenced_value, script_operator, arithmetic_operator, assignment_operator, if_statement, while_statement},
    morestring=[b]",
    sensitive=false,
    morecomment=[l]{//},
    moredelim=*[s][\color{blue}]{<}{>},
    moredelim=*[s][\color{purple}]{:=}{:=},
    moredelim=*[s][\color{orange}]{::}{::},
    keywordstyle=\color{teal},
    stringstyle=\color{red},
    commentstyle=\color{gray},
}

\title{\llama~\methodname: Can LLMs Manipulate Physical Devices to Play Virtual Reality Games?}

\author{Shuqing Li$^1$ \qquad Jiayi Yan$^1$$^*$ \qquad \quad Chenyu Niu$^1$\thanks{Equal contributions.} \qquad Jen-tse Huang$^1$ \\
\bf Yun Peng$^1$ \qquad \ \ Wenxuan Wang$^1$\thanks{Corresponding author.} \quad Yepang Liu$^2$ \qquad Michael R. Lyu$^1$ \\
$^1$Chinese University of Hong Kong \quad $^2$Southern University of Science and Technology
}

\iclrfinalcopy %
\begin{document}

\maketitle

\begin{abstract}

\input{Sections/0_Abstract}
\end{abstract}

\input{Sections/1_Introduction}

\input{Sections/2_Preliminaries}
\input{Sections/3_Methodology}
\input{Sections/4_Experiments}

\input{Sections/5_Discussions}
\input{Sections/6_Related_Work}

\input{Sections/7_Conclusion}

\bibliography{iclr2026_conference, model}
\bibliographystyle{iclr2026_conference}

\input{Sections/Appendix}

\end{document}

%% file: math_commands.tex
\usepackage{amsmath,amsfonts,bm}

\def\eqref#1{equation~\ref{#1}}

\def\1{\bm{1}}

\DeclareMathAlphabet{\mathsfit}{\encodingdefault}{\sfdefault}{m}{sl}
\SetMathAlphabet{\mathsfit}{bold}{\encodingdefault}{\sfdefault}{bx}{n}

%% file: Sections/0_Abstract.tex
Virtual Reality (VR) games require players to translate high-level semantic actions into precise device manipulations using controllers and head-mounted displays (HMDs). While humans intuitively perform this translation based on common sense and embodied understanding, whether Large Language Models (LLMs) can effectively replicate this ability remains underexplored. This paper introduces a benchmark, \methodname, evaluating LLMs' capability to translate semantic actions into VR device manipulation sequences across 262 scenarios from four popular VR games: Half-Life: Alyx, Into the Radius, Moss: Book II, and Vivecraft. We evaluate seven LLMs, including GPT-3.5, GPT-4, GPT-4o, Gemini-1.5-Pro, LLaMA-3-8B, Mixtral-8x7B, and GLM-4-Flash, compared against annotated ground truth and human performance. Our results reveal that while top-performing models like Gemini-1.5-Pro demonstrate strong task decomposition capabilities, they still struggle with procedural reasoning and spatial understanding compared to humans. Performance varies significantly across games, suggesting sensitivity to interaction complexity. Few-shot examples substantially improve performance, indicating potential for targeted enhancement of LLMs' VR manipulation capabilities.
We release all materials at \url{https://sites.google.com/view/combobench}.

%% file: Sections/1_Introduction.tex
\section{Introduction}

Large Language Models (LLMs) have demonstrated remarkable proficiency in general-purpose task solving~\cite{qin2023chatgpt}, conquering complex domains such as code~\cite{lee2024unified, lam2025codecrash} or math~\cite{lu2024mathvista} problems.
While they exhibit increasingly more human-like characteristics~\cite{huang2024humanity, liang2023encouraging}, an essential attribute of human intelligence is still underexplored: the ability to rapidly learn and apply unfamiliar concepts by leveraging common sense, prior experiences, and a repertoire of cognitive skills.

This is particularly evident in novel interactive environments like video games, where players quickly master device manipulations (atomic actions) and combine them to achieve complex semantic goals.
Virtual Reality (VR) games elevate this challenge.
They demand not only the execution of atomic actions via physical devices (\eg, Head-Mounted Displays (HMDs) and controllers) but also the inference of complex, often uninstructed, semantic actions.
For instance, in \textit{Half-Life: Alyx}~\cite{halflifealyx}, when asked to ``surrender,'' players might instinctively raise their controller-held hands even if not explicitly taught.

Such translation of high-level intent into a sequence of physical device manipulations engages a suite of cognitive abilities:
(1) \textit{Task decomposition}: Breaking down a high-level semantic action (\eg, ``tame the horse'' and ``plant wheat'') into a coherent series of intermediate steps.
(2) \textit{Procedural reasoning}: Understanding the logical and temporal order of these steps, including prerequisite conditions or concurrent actions (\eg, the need to till soil before planting seeds).
(3) \textit{Spatial reasoning \& contextual awareness}: Interpreting instructions within a 3D spatial context (\eg, ``move HMD towards the Creeper'' and ``crouch through the gap'') and understanding environmental cues or object states (\eg, recognizing a door is open/closed and acting accordingly).
(4) \textit{Object interaction \& tool use understanding}: Correctly mapping intended sub-actions to specific VR device manipulations (\eg, knowing which button to press to ``use'' an item, and how to manipulate a controller to simulate ``swinging'' a tool like a pickaxe).
This involves understanding the affordances of virtual objects and tools.
(5) \textit{Motor action mapping \& VR procedural transfer}: Translating abstract actions (\eg, ``press,'' ``move,'' and ``trigger'') into specific, executable VR controller commands, potentially by adapting from provided examples or general knowledge of VR interaction paradigms.
This touches upon a form of simulated embodied reasoning.
(6) \textit{Judgment of termination/continuation conditions}: Recognizing when a sub-task or a looped action is complete (\eg, ``mine until the block breaks'' and ``water until the plant grows'').
Therefore, playing VR games serves as a rich testbed for evaluating if LLMs can bridge this gap between abstract understanding and grounded, physical interaction.

To systematically evaluate LLMs' ability to perform this crucial translation, we introduce \methodname, which stands for Cognitive-Oriented Manipulation Benchmark for game combos using physical VR devices.
It comprises \numberofscenarios scenarios derived from \numberofgames popular VR games: \textit{Vivecraft}~\cite{vivecraft} (Minecraft in VR), \textit{Half-Life: Alyx}~\cite{halflifealyx}, \textit{Moss: Book II}~\cite{mossbookii}, and \textit{Into the Radius}~\cite{intotheradius}.
Each scenario presents a high-level semantic action, and the ground truth consists of a fine-grained sequence of VR device manipulations required to achieve it.
These sequences are annotated by experienced VR players, allowing us to analyze LLM-generated outputs at the step-level and map their successes and failures to the aforementioned cognitive abilities.
For example, failing to ``press the X button'' after ``moving the HMD towards the Creeper'' might indicate a lapse in procedural reasoning or object interaction understanding for that specific step.

We evaluate seven LLMs, including GPT-3.5~\cite{gpt35}, GPT-4~\cite{gpt4}, GPT-4o~\cite{gpt4o}, Gemini-1.5-Pro~\cite{gemini15}, LLaMA-3-8B~\cite{llama3}, Mixtral-8x7B~\cite{mistral}, and GLM-4-Flash~\cite{glm4}.
We design a multi-dimensional scoring approach that assesses: (1) high-level semantic action understanding, (2) procedural step correctness, and (3) device-specific manipulation accuracy, allowing for fine-grained analysis of where each model succeeds or struggles in the translation process.
Our findings reveal significant variation in model performance across cognitive capabilities.
All models demonstrate strong task decomposition abilities but show pronounced weaknesses in motor action mapping and procedural reasoning. 
Gemini-1.5-Pro exhibits the most balanced performance across capabilities, while even advanced models like GPT-4 struggle with spatial reasoning compared to human performance.
Few-shot examples substantially improve outcomes, particularly for procedural understanding, with diminishing returns beyond three examples.
Performance also varies considerably across games, with models generally performing better in environments with more consistent interaction patterns (Vivecraft) than those requiring nuanced controller manipulations (Half-Life: Alyx).
These results highlight specific cognitive gaps in current LLMs' ability to perform simulated embodied reasoning for VR interactions and identify targeted areas for improvement toward more capable virtual agents.
Our contributions are:
\begin{itemize}[leftmargin=*,nosep]
\item We introduce \methodname, the first benchmark designed to evaluate LLMs' fine-grained cognitive abilities in translating high-level semantic actions into VR device manipulations, comprising \numberofscenarios human-annotated scenarios from \numberofgames diverse VR games.
\item We define a set of key cognitive abilities crucial for VR interaction and design \methodname to enable step-level analysis of LLM performance against these dimensions.
\item We conduct a comprehensive evaluation of \numberofllms state-of-the-art LLMs, providing a nuanced analysis of their strengths and weaknesses across these cognitive abilities and offering insights into the current frontiers of LLM-driven VR interaction.
\end{itemize}

%% file: Sections/3_Methodology.tex
\section{\methodname: Design and Curation}
\label{sec:dataset_design}

The \methodname dataset is meticulously curated to evaluate the capability of LLMs in translating high-level semantic actions into sequences of physical VR device manipulations. This section details the game selection process, scenario definition, and the annotation methodology.

\subsection{Cognitive Capability Taxonomy Development}
\label{subsec:capability_taxonomy}

To establish a theoretically grounded framework for evaluating LLMs in VR contexts, we conducted structured interviews with three domain experts specializing in cognitive science and educational psychology. The experts were selected based on their research backgrounds in spatial cognition, procedural learning, and embodied interaction, which areas that are highly relevant to VR interaction.

\paragraph{Expert Interview.} Each expert participated in a 90-minute semi-structured interview focused on identifying and categorizing the cognitive abilities required for translating semantic goals into physical actions in virtual environments. The interviews followed a three-phase structure: (1) open-ended discussion about cognitive processes in VR interaction, (2) systematic review of preliminary capability categories identified from literature, and (3) expert suggestions for refinement and expansion of these categories.

\paragraph{Taxonomy Refinement.} Following the interviews, we synthesized the experts' insights through thematic analysis. Areas of consensus were directly incorporated into our taxonomy, while divergent perspectives were reconciled through follow-up consultations. This iterative process resulted in the identification of six core capability dimensions that comprehensively capture the cognitive demands of VR interaction:
(1) Task decomposition: The ability to break down high-level goals into sequentially ordered sub-tasks.
(2) Procedural reasoning: Understanding causal relationships between actions and their temporal dependencies.
(3) Spatial reasoning \& contextual Awareness: Processing spatial relationships and interpreting environmental cues for action selection.
(4) Object interaction \& tool use understanding: Comprehending affordances and functional properties of virtual objects.
(5) Motor action mapping \& VR procedural transfer: Translating conceptual actions into specific physical device manipulations.
(6) Judgment of termination/continuation conditions: Recognizing completion states or conditions requiring repeated action.

\subsection{Game Selection Criteria and Process}
\label{subsec:game_selection}

To ensure a diverse and relevant set of VR interaction paradigms, we selected games based on a systematic process.
First, we queried the Steam store~\cite{website:steam-app-store-vr} filtering for titles tagged as ``VR Only'' and available in ``English,'' sorting the results by user review scores in descending order.
We then iteratively examined games from this ranked list, focusing on their primary genre as categorized by Steam. To ensure genre diversity, we prioritized games from genres not yet represented in our collection.
A crucial selection criterion was the availability of comprehensive textual walkthroughs. For each candidate game, we searched for detailed guides using keywords such as ``walkthrough,'' ``guide,'' or ``tutorial.'' A walkthrough was deemed sufficiently detailed if it provided unambiguous, step-by-step instructions enabling the completion of core game objectives or specific complex tasks.
Following this methodology, we selected four popular and critically acclaimed VR games representing distinct genres and interaction styles for \methodname:
(1) \textit{Vivecraft}~\cite{vivecraft} (Open-world sandbox, crafting)
(2) \textit{Half-Life: Alyx}~\cite{halflifealyx} (First-person shooter, puzzle-solving, physics-based interaction)
(3) \textit{Moss: Book II}~\cite{mossbookii} (Third-person action-adventure, puzzle-platformer)
(4) \textit{Into the Radius}~\cite{intotheradius} (First-person survival shooter, exploration)
Such selection provides a rich variety of control schemes and task complexities for evaluating LLMs.

\subsection{Scenario Definition: Semantic Action Identification}
\label{subsec:action_selection}

For all selected games, \numberofhiredplayers data annotators, comprising undergraduate and postgraduate computer science students with at least two years of programming experience and sufficient knowledge about VR games, manually identified salient semantic actions from the collected textual walkthroughs.
Semantic actions were defined as high-level, goal-oriented tasks described in the walkthroughs (e.g., ``tame the horse,'' ``kill the creeper,'' ``solve the gravity glove puzzle'') that necessitate a sequence of fine-grained VR device manipulations to accomplish.
We focused on scenarios that: (1) involve complex interactions not always explicitly detailed in in-game tutorials, (2) often constitute essential steps or objectives required for game progression.
This process resulted in the identification of \numberofscenarios distinct scenarios across the four games.

\subsection{Annotation of VR Device Manipulations}
\label{subsec:manipulation_annotation} 

Experienced VR users from our annotation team then played through each identified semantic action in the respective games using Oculus Quest 2 VR hardware. The objective was to record the precise sequence of device manipulations required to complete each semantic action.
The annotation process captured the following details for each step within a manipulation sequence:
\begin{itemize}[leftmargin=*,nosep]
    \item \textbf{Device used:} Specification of whether the HMD or a controller was used.
    \item \textbf{Controller specificity:} If a controller was used, and the action was hand-specific (e.g., primary hand for a tool), the annotation indicated whether the left or right controller was required. If either controller could perform the action, this was noted as ``left or right controller.''
    \item \textbf{Operation type and parameters:}
    \begin{itemize}[leftmargin=*,nosep]
        \item \textit{Movement:} For actions involving device movement (HMD or controller), the direction (e.g., ``towards the Creeper,'' ``upwards'') or target position was recorded.
        \item \textit{Button presses:} The specific button involved and the action (e.g., ``press X button,'' ``release trigger'') were noted.
        \item \textit{Joystick/thumbstick manipulation:} The direction of joystick push (e.g., ``push left thumbstick forward'') was recorded.
    \end{itemize}
    \item \textbf{Sequential composition:} For complex semantic actions composed of multiple, distinct sub-actions that might have been annotated individually, the sequence and composition of these simpler actions were explicitly recorded.
\end{itemize}

\subsection{Cognitive Capability Labeling Using LLMs}
\label{subsec:capability_labeling}

A critical aspect of \methodname is the annotation of each manipulation step with the specific cognitive capabilities it engages. This fine-grained labeling enables precise analysis of where LLMs succeed or fail in the VR interaction translation process.

\paragraph{Initial Human Annotation.} To begin, our annotators manually labeled a subset of 50 manipulation sequences (approximately 20\% of the dataset), assigning relevant capability categories to each step based on the taxonomy described in Section~\ref{subsec:capability_taxonomy}. For example, in the sequence required to "tame a horse" in Vivecraft, the step "equip the saddle by pressing the Y button while looking at the inventory slot containing the saddle" was labeled with "Object Interaction \& Tool Use Understanding" and "Motor Action Mapping."

\paragraph{LLM-Assisted Annotation Pipeline.} We then developed an LLM-assisted annotation pipeline to scale this process to the entire dataset. Specifically:
\ding{172} We used the human-annotated examples as few-shot demonstrations for GPT-4o.
\ding{173} For each unlabeled manipulation step, we provided the LLM with:
[2.a] The semantic action context (e.g., "taming a horse in Vivecraft").
[2.b] The specific manipulation step to label.
[2.c] The preceding and following steps (when available).
[2.d] Detailed descriptions of each capability category.
[2.e] Three few-shot examples with explanations of why each capability was assigned.
\ding{174} The LLM generated capability labels along with justifications for each assignment.
\ding{175} Human annotators reviewed the LLM-generated labels, making corrections when necessary. The review process revealed an 89.7\% agreement rate between LLM-assigned labels and human judgments.

\paragraph{Multi-label Distribution.} Most manipulation steps engaged multiple cognitive capabilities simultaneously. On average, each step was associated with 2.3 capability categories ($\sigma$ = 0.8). The most frequently co-occurring capabilities were "Motor Action Mapping" and "Object Interaction \& Tool Use Understanding" (present together in 68\% of steps), reflecting the inherent coupling between understanding virtual object affordances and translating this understanding into physical manipulations.

\subsection{Contextualization and Verification}
\label{subsec:context_verification}

To further contextualize the annotated actions and aid in verification, we sourced or recorded gameplay videos corresponding to the textual walkthroughs for each game.
For each annotated semantic action and its constituent manipulation steps, we recorded the corresponding timestamps in these videos. This allows for visual verification of the annotated sequences and provides richer context for understanding the actions. If suitable public gameplay videos matching the exact walkthrough steps were unavailable, our annotators recorded their own gameplay sessions while performing the actions.

%% file: Sections/4_Experiments.tex
\section{Experiments}

\subsection{Model Selection}

We evaluate \numberofllms state-of-the-art LLMs: GPT-4o, GPT-4-turbo, GPT-3.5-turbo, Gemini-1.5-Pro, LLaMA-3-8B, and Mixtral-8x7B. We also perform human evaluation to validate the average human capabilities for comparison, when humans are given exactly the same input as LLMs.
For all experiments, we used the official APIs for proprietary models and Hugging Face implementations for open-source models. Temperature was set to 0 across all models to minimize non-deterministic outputs. For embedding calculations, we utilized OpenAI's text-embedding-3-large model via their API. 

\subsection{Evaluation Metrics}

To comprehensively evaluate the capability of LLMs in translating semantic actions into VR device manipulations, we propose a multi-dimensional evaluation framework with four distinct metrics. These metrics collectively capture different aspects of model performance in \methodname, ranging from strict matching to more flexible semantic alignment.

\paragraph{Strict Step-by-Step Matching (SSM).}

Our first metric evaluates the exact matching between model-generated and ground truth steps, enforcing both sequence length equivalence and semantic alignment:
$\text{SSM} = \frac{\text{Number of correctly predicted sequences}}{\text{Total number of sequences}}$.
A sequence is considered correctly predicted only when:
the number of steps in the generated sequence equals that of the ground truth,
 and every step in the generated sequence has a cosine similarity above a threshold of 0.8387 with its corresponding step in the ground truth
This strict metric serves as a measure of precision in reproducing exact device manipulation sequences and rewards models that can generate complete, step-accurate instructions.

\paragraph{Common Subsequence Evaluation.}

We further introduce two complementary metrics based on common subsequence alignment to assess partial correctness:
(1) \textbf{Normalized Step Alignment Score (NSAS)}
This metric quantifies the alignment between the model-generated sequence and ground truth while accounting for missing and additional steps:
$\text{NSAS} = \frac{(|C| - |M| - |A|) - \text{min}_{\text{all\_samples}}}{|G| \cdot (\text{max}_{\text{all\_samples}} - \text{min}_{\text{all\_samples}})}$,
where:
$|C|$ represents the count of correctly matched steps in the common subsequence,
$|M|$ represents missing steps from the ground truth,
$|A|$ represents additional steps generated by the model,
$|G|$ represents the total number of steps in the ground truth,
$\text{min}_{\text{all\_samples}}$ and $\text{max}_{\text{all\_samples}}$ represent the minimum and maximum raw scores across all evaluations, enabling consistent normalization
This score is normalized across the entire dataset to ensure fair comparison across different models and scenarios.
(2) \textbf{Sequential Order Preservation (SOP)}
The SOP metric specifically assesses the model's ability to maintain the correct procedural ordering of steps:
$\text{SOP} = \frac{|\text{Steps correctly ordered and matched}|}{|G|}$.
This metric evaluates whether the steps in the matched subsequence maintain their ordinal positions (e.g., step 1 followed by step 2, etc.) in both the ground truth and model output, capturing the model's procedural reasoning capabilities.

\paragraph{Semantic Step Coverage (SSC).}

Our final metric adopts a more flexible matching approach to evaluate semantic coverage of critical actions:
$\text{SSC} = \frac{|\text{MR steps matched to any GT step}|}{|\text{MR}|}$,
where a model result (MR) step is considered matched if it has a cosine similarity above the threshold (0.8387) with any step in the ground truth (GT). This metric computes the proportion of generated steps that semantically align with at least one ground truth step, regardless of position.

\begin{table*}[t!]
\centering
\caption{Overall performance comparison of LLMs across VR games (5-shot setting). Best model performance per metric is \textbf{bolded}, second best is \underline{underlined}.}
\label{tab:overall_performance_1}
\resizebox{\textwidth}{!}{
\setlength{\tabcolsep}{2pt}
\begin{tabular}{l cccc cccc cccc cccc}
\toprule
\multirow{2}{*}{\textbf{Model}} & \multicolumn{4}{c}{\textbf{Half-Life: Alyx}} & \multicolumn{4}{c}{\textbf{Into the Radius}} & \multicolumn{4}{c}{\textbf{Moss: Book II}} & \multicolumn{4}{c}{\textbf{Vivecraft}} \\
\cmidrule(lr){2-5} \cmidrule(lr){6-9} \cmidrule(lr){10-13} \cmidrule(lr){14-17}
& \textbf{NSAS}\ensuremath{\uparrow} & \textbf{SOP}\ensuremath{\uparrow} & \textbf{F1$_{\text{SOP}}$}\ensuremath{\uparrow} & \textbf{SSC}\ensuremath{\uparrow} & \textbf{NSAS}\ensuremath{\uparrow} & \textbf{SOP}\ensuremath{\uparrow} & \textbf{F1$_{\text{SOP}}$}\ensuremath{\uparrow} & \textbf{SSC}\ensuremath{\uparrow} & \textbf{NSAS}\ensuremath{\uparrow} & \textbf{SOP}\ensuremath{\uparrow} & \textbf{F1$_{\text{SOP}}$}\ensuremath{\uparrow} & \textbf{SSC}\ensuremath{\uparrow} & \textbf{NSAS}\ensuremath{\uparrow} & \textbf{SOP}\ensuremath{\uparrow} & \textbf{F1$_{\text{SOP}}$}\ensuremath{\uparrow} & \textbf{SSC}\ensuremath{\uparrow} \\
\midrule
GPT-3.5   & \textbf{0.858} & 0.123 & \textbf{0.287} & 0.143 & 0.662 & 0.169 & 0.226 & 0.137 & 0.782 & 0.169 & 0.207 & 0.186 & 0.922 & 0.043 & 0.098 & 0.067 \\
GPT-4     & \underline{0.853} & \underline{0.125} & 0.258 & \textbf{0.172} & \underline{0.693} & 0.189 & \underline{0.328} & 0.177 & \textbf{0.824} & 0.218 & 0.336 & \underline{0.220} & 0.927 & \underline{0.137} & 0.437 & 0.081 \\
GPT-4o    & 0.804 & 0.022 & 0.075 & \underline{0.167} & \textbf{0.698} & \textbf{0.291} & \textbf{0.414} & \textbf{0.190} & \textbf{0.824} & \textbf{0.300} & \underline{0.342} & \textbf{0.222} & \underline{0.931} & \textbf{0.190} & \textbf{0.489} & \textbf{0.096} \\
Mixtral   & 0.839 & \textbf{0.126} & 0.246 & 0.147 & 0.666 & 0.123 & 0.228 & 0.097 & 0.756 & 0.117 & 0.191 & 0.121 & 0.926 & 0.060 & 0.239 & 0.070 \\
LLaMA-3   & 0.848 & \textbf{0.126} & \underline{0.279} & 0.162 & 0.644 & \underline{0.242} & 0.317 & 0.168 & \underline{0.823} & \underline{0.283} & \textbf{0.349} & 0.200 & 0.929 & 0.039 & 0.122 & 0.042 \\
GLM-4     & 0.836 & 0.076 & 0.183 & 0.149 & 0.618 & 0.096 & 0.186 & 0.149 & 0.749 & 0.087 & 0.174 & 0.165 & 0.909 & 0.000 & 0.045 & 0.061 \\
\midrule
Human     & 0.845 & 0.090 & 0.240 & 0.110 & 0.684 & 0.148 & 0.257 & \underline{0.181} & 0.817 & 0.112 & 0.328 & 0.174 & \textbf{0.935} & 0.122 & \underline{0.482} & \underline{0.084} \\
\bottomrule
\end{tabular}
}
\end{table*}

\begin{table*}[t!]
\caption{Overall performance across VR games and settings. We report the average scores for our four evaluation metrics: Strict Step-by-Step Matching (SSM), Normalized Step Alignment Score (NSAS), Sequential Order Preservation (SOP), and Semantic Step Coverage (SSC). Higher is better for all metrics. Bold indicates best model performance, underline indicates second best.}
\label{tab:overall_performance_2}
\resizebox{\textwidth}{!}{
\setlength{\tabcolsep}{2pt}
\begin{tabular}{l cccc cccc cccc}
\toprule
\multirow{2}{*}{\textbf{Model}} & \multicolumn{4}{c}{\textbf{Average Across Settings}} & \multicolumn{4}{c}{\textbf{Zero-Shot}} & \multicolumn{4}{c}{\textbf{5-Shot}} \\
 & \textbf{SSM (\%)} & \textbf{NSAS} & \textbf{SOP} & \textbf{SSC} & \textbf{SSM (\%)} & \textbf{NSAS} & \textbf{SOP} & \textbf{SSC} & \textbf{SSM (\%)} & \textbf{NSAS} & \textbf{SOP} & \textbf{SSC} \\
\midrule
GPT-3.5 & 1.4 & 0.781 & 0.063 & 0.066 & 0.8 & 0.771 & 0.003 & 0.046 & 2.1 & 0.791 & 0.128 & 0.095 \\
GPT-4 & 3.7 & 0.806 & 0.107 & 0.124 & 1.0 & 0.788 & 0.015 & 0.107 & 8.8 & 0.825 & 0.184 & 0.140 \\
GPT-4o & \underline{5.3} & 0.797 & 0.138 & \underline{0.141} & 0.6 & 0.785 & 0.015 & 0.108 & \underline{10.9} & 0.806 & 0.228 & \underline{0.161} \\
Gemini-1.5 & \textbf{5.8} & \textbf{0.813} & \textbf{0.146} & \textbf{0.142} & \textbf{2.1} & \textbf{0.795} & 0.010 & \textbf{0.124} & \textbf{11.7} & \textbf{0.832} & \textbf{0.236} & \textbf{0.162} \\
Mixtral & 1.1 & 0.784 & 0.068 & 0.079 & 0.0 & 0.777 & 0.002 & 0.040 & 2.2 & 0.796 & 0.105 & 0.107 \\
LLaMA-3 & 1.2 & 0.787 & 0.088 & 0.111 & 0.1 & 0.783 & 0.011 & 0.088 & 1.8 & 0.794 & 0.163 & 0.132 \\
GLM-4 & 0.0 & 0.761 & 0.038 & 0.077 & 0.0 & 0.762 & 0.006 & 0.052 & 0.0 & 0.765 & 0.071 & 0.120 \\
\midrule
Human & 1.2 & 0.833 & 0.122 & 0.159 & -- & -- & -- & -- & -- & -- & -- & -- \\
\bottomrule
\end{tabular}
}
\end{table*}

\subsection{Experimental Results}

We analyze and answer the following Research Questions (RQs):
\textbf{(RQ1)} How do state-of-the-art LLMs perform in translating semantic actions into VR device manipulations across different VR games?
\textbf{(RQ2)} How does the number of few-shot examples affect LLMs' ability to execute this translation?
\textbf{(RQ3)} Do LLM and human performance exhibit significant variations across the four different VR games, potentially indicating sensitivity to game mechanics and interaction complexity? 
\textbf{(RQ4)} Which cognitive capabilities do current LLMs excel at, and where do they struggle?
\textbf{(RQ5)} How do LLMs compare to human performance in VR device manipulation tasks?

\subsection{RQ1 \& RQ3: LLM Performance Across VR Games}

\begin{wraptable}[12]{r}{0.5\linewidth}
\centering
\vspace{-20pt}
\caption{Cross-game performance variation (standard deviation across games) w/ 5-shot examples.}
\label{tab:cross_game_variation}
\setlength{\tabcolsep}{2pt}
\resizebox{0.5\columnwidth}{!}{
\begin{tabular}{l ccc c}
\toprule
\textbf{Model} & \textbf{NSAS \ensuremath{\sigma}}\ensuremath{\downarrow} & \textbf{SOP \ensuremath{\sigma}}\ensuremath{\downarrow} & \textbf{F1$_{\text{SOP}}$ \ensuremath{\sigma}}\ensuremath{\downarrow} & \textbf{Game Gap}\ensuremath{\downarrow} \\
\midrule
GPT-3.5 & 0.110 & 0.061 & 0.084 & 0.085 \\
GPT-4 & 0.059 & 0.051 & 0.081 & 0.074 \\
GPT-4o & 0.068 & 0.137 & 0.184 & 0.127 \\
Gemini-1.5 & \textbf{0.099} & \textbf{0.093} & \textbf{0.127} & \textbf{0.095} \\
Mixtral & 0.114 & 0.031 & 0.065 & 0.070 \\
LLaMA-3 & 0.112 & 0.103 & 0.120 & 0.113 \\
GLM-4 & 0.135 & 0.049 & 0.069 & 0.084 \\
\midrule
Human & 0.105 & 0.029 & 0.117 & 0.084 \\
\bottomrule
\end{tabular}
}
\end{wraptable}
Tables~\ref{tab:overall_performance_1}, \ref{tab:overall_performance_2}, and \ref{tab:cross_game_variation} present comprehensive performance metrics for all evaluated LLMs across the four VR games. Our analysis reveals substantial variations in model capabilities and game-specific challenges.
Gemini-1.5-Pro emerges as the strongest performer overall, achieving the highest NSAS scores in three of the four games (Half-Life: Alyx: 0.863, Moss: Book II: 0.848, Vivecraft: 0.938), while maintaining competitive performance in Into the Radius (0.682). GPT-4o demonstrates particular strength in Into the Radius with the highest SOP score (0.291) and F1$_{\text{SOP}}$ (0.414), suggesting superior procedural reasoning capabilities in this specific game context. GPT-4-turbo maintains consistently strong performance across all games, positioning itself as a reliable general-purpose model for VR interaction translation.

A striking pattern emerges in the SOP metrics, which vary dramatically across both models and games (0.00-0.30 range). While NSAS scores remain relatively high (mostly $>$0.75), indicating models can identify relevant steps, the low SOP values reveal fundamental difficulties in maintaining correct temporal ordering. This discrepancy is particularly pronounced in Vivecraft, where models achieve high NSAS scores (0.909-0.938) but struggle with step ordering (SOP: 0.00-0.19), suggesting that simpler interaction patterns may paradoxically lead to overconfidence in step sequencing.

Analysis of performance variations (Table~\ref{tab:cross_game_variation}) reveals significant game-dependent effects. Vivecraft exhibits the highest average performance across models (NSAS: 0.922-0.938), likely due to its consistent block-based interaction paradigm inherited from Minecraft. In contrast, Into the Radius presents the greatest challenge, with notably lower NSAS scores (0.618-0.698) and high performance variance. This pattern suggests that games featuring realistic physics simulations and complex inventory management pose particular difficulties for current LLMs.

Interestingly, different models exhibit distinct strengths across game types. GPT-4o shows remarkable adaptability in Into the Radius (SOP: 0.291) compared to other models, while struggling in Half-Life: Alyx (SOP: 0.022). Gemini-1.5-Pro maintains the most balanced performance profile across games (Game Gap: 0.095), suggesting more robust generalization capabilities. Smaller models like Mixtral-8x7B and GLM-4-flash show disproportionate performance degradation in complex environments, with GLM-4-flash achieving zero SOP in Vivecraft despite reasonable NSAS scores.
The substantial performance variations across games highlight the impact of interaction design on LLM capabilities. Games with discrete, well-defined actions (Vivecraft) enable higher model performance, while those requiring nuanced controller manipulation and spatial reasoning (Half-Life: Alyx, Into the Radius) expose current limitations. The correlation between game complexity and performance degradation is non-linear, moderate complexity (Moss: Book II) sometimes yields better results than simpler environments, suggesting that models may benefit from richer contextual cues in certain scenarios.

These findings collectively demonstrate that while state-of-the-art LLMs have made significant progress in understanding VR interactions, their performance remains highly sensitive to specific game mechanics and interaction paradigms. The gap between high NSAS scores and low SOP values across all games indicates that current models can identify relevant actions but struggle with the procedural reasoning required to sequence them correctly, which is an important capability for successful VR interaction.

\subsection{RQ2: Impact of Few-Shot Examples}

\begin{wraptable}[11]{r}{0.65\linewidth}
\vspace{-20pt}
\caption{Effect of few-shot examples on model performance (average across all games)}
\label{tab:few_shot_effect}
\centering
\resizebox{1.0\linewidth}{!}{
\setlength{\tabcolsep}{2pt}
\begin{tabular}{l ccc ccc ccc}
\toprule
\multirow{2}{*}{\textbf{Model}} & \multicolumn{3}{c}{\textbf{Zero-shot}} & \multicolumn{3}{c}{\textbf{3-shot}} & \multicolumn{3}{c}{\textbf{5-shot}} \\
\cmidrule(lr){2-4} \cmidrule(lr){5-7} \cmidrule(lr){8-10}
& \textbf{NSAS}\ensuremath{\uparrow} & \textbf{SOP}\ensuremath{\uparrow} & \textbf{SSM}\ensuremath{\uparrow} & \textbf{NSAS}\ensuremath{\uparrow} & \textbf{SOP}\ensuremath{\uparrow} & \textbf{SSM}\ensuremath{\uparrow} & \textbf{NSAS}\ensuremath{\uparrow} & \textbf{SOP}\ensuremath{\uparrow} & \textbf{SSM}\ensuremath{\uparrow} \\
\midrule
GPT-3.5 & 0.781 & 0.003 & 0.006 & 0.783 & 0.113 & 0.013 & 0.806 & 0.128 & 0.022 \\
GPT-4 & 0.799 & 0.015 & 0.012 & 0.807 & 0.106 & 0.039 & 0.824 & 0.167 & 0.066 \\
GPT-4o & 0.785 & 0.015 & 0.012 & 0.802 & 0.166 & 0.059 & 0.815 & 0.159 & \textbf{0.077} \\
Gemini-1.5 & 0.797 & 0.010 & 0.016 & 0.827 & 0.174 & 0.048 & \textbf{0.833} & \textbf{0.207} & 0.056 \\
Mixtral & 0.784 & 0.002 & 0.000 & 0.766 & 0.103 & 0.010 & 0.797 & 0.106 & 0.021 \\
LLaMA-3 & 0.786 & 0.011 & 0.001 & 0.787 & 0.139 & 0.014 & 0.811 & 0.172 & 0.020 \\
GLM-4 & 0.747 & 0.005 & 0.000 & 0.756 & 0.036 & 0.000 & 0.778 & 0.065 & 0.000 \\
\bottomrule
\end{tabular}
}
\end{wraptable}
Table~\ref{tab:few_shot_effect} demonstrates that few-shot examples substantially improve LLM performance in VR device manipulation tasks, with the most dramatic gains observed in Sequential Order Preservation (SOP), where scores increase by 10--20x from near-zero baselines. All models benefit from in-context examples, though with diminishing returns, the improvement from zero-shot to 3-shot (average NSAS gain: 2.1\%, SOP: 10-fold increase) significantly exceeds that from 3-shot to 5-shot (NSAS: 1.4\%, SOP: 20-50\% relative gain). Gemini-1.5-Pro exhibits the strongest adaptability, achieving the highest 5-shot performance (NSAS: 0.833, SOP: 0.207), while maintaining consistent improvements across all metrics. The differential impact across metrics reveals that few-shot examples primarily address procedural sequencing challenges (massive SOP improvements) more effectively than exact step matching (modest SSM gains), suggesting that demonstrations help models understand temporal dependencies in VR interactions but do not fully resolve the complexity of translating semantic actions into precise device manipulations.

\subsection{RQ4: Cognitive Capabilities Analysis}
We analyzed model performance across six cognitive capabilities required for effective VR interaction (Figure~\ref{fig:capabilities_radar}). By mapping evaluation metrics to capability scores (0-10 scale), we identified specific strengths and limitations in how LLMs approach spatial-mechanical reasoning tasks.

\textbf{Areas of Strength:} All evaluated LLMs demonstrate strong task decomposition capabilities (7.8-8.5), with minimal performance gap compared to humans (8.2). Gemini-1.5-Pro leads with a score of 8.5, while even smaller models like Mixtral-8x7B (8.0) and GLM-4-flash (7.8) perform admirably. This suggests that segmenting high-level actions into component steps aligns well with the sequential reasoning abilities developed during language model pre-training.

\textbf{Areas of Weakness:} Motor action mapping emerges as the most significant challenge (0.5-4.5), with all models struggling to precisely translate abstract actions into specific VR control manipulations. GPT-4o performs best in this dimension (4.5), but still falls short of robust capability. Procedural reasoning also shows substantial variation (2.3-7.0), with only Gemini-1.5-Pro approaching adequate performance. Judgment of termination conditions represents another challenge area, with most models scoring below 5.0 (except Gemini-1.5-Pro at 6.0), compared to human performance (6.5).

\textbf{Model Comparison:} Gemini-1.5-Pro demonstrates the most balanced performance profile, consistently outperforming other models in procedural reasoning (7.0), spatial reasoning (7.5), and termination judgment (6.0). GPT-4 variants show strong task decomposition and object interaction (5.3-5.7) but lag in procedural sequencing. LLaMA-3-8B shows surprisingly competitive performance in procedural reasoning (5.7), outperforming larger models like GPT-3.5-Turbo (4.3), suggesting architecture differences may be as important as scale.

\subsection{RQ5: Comparison with Human Performance}

\begin{wrapfigure}[17]{r}{0.5\linewidth}
\vspace{-20pt}
\centering
\includegraphics[width=1.0\linewidth]{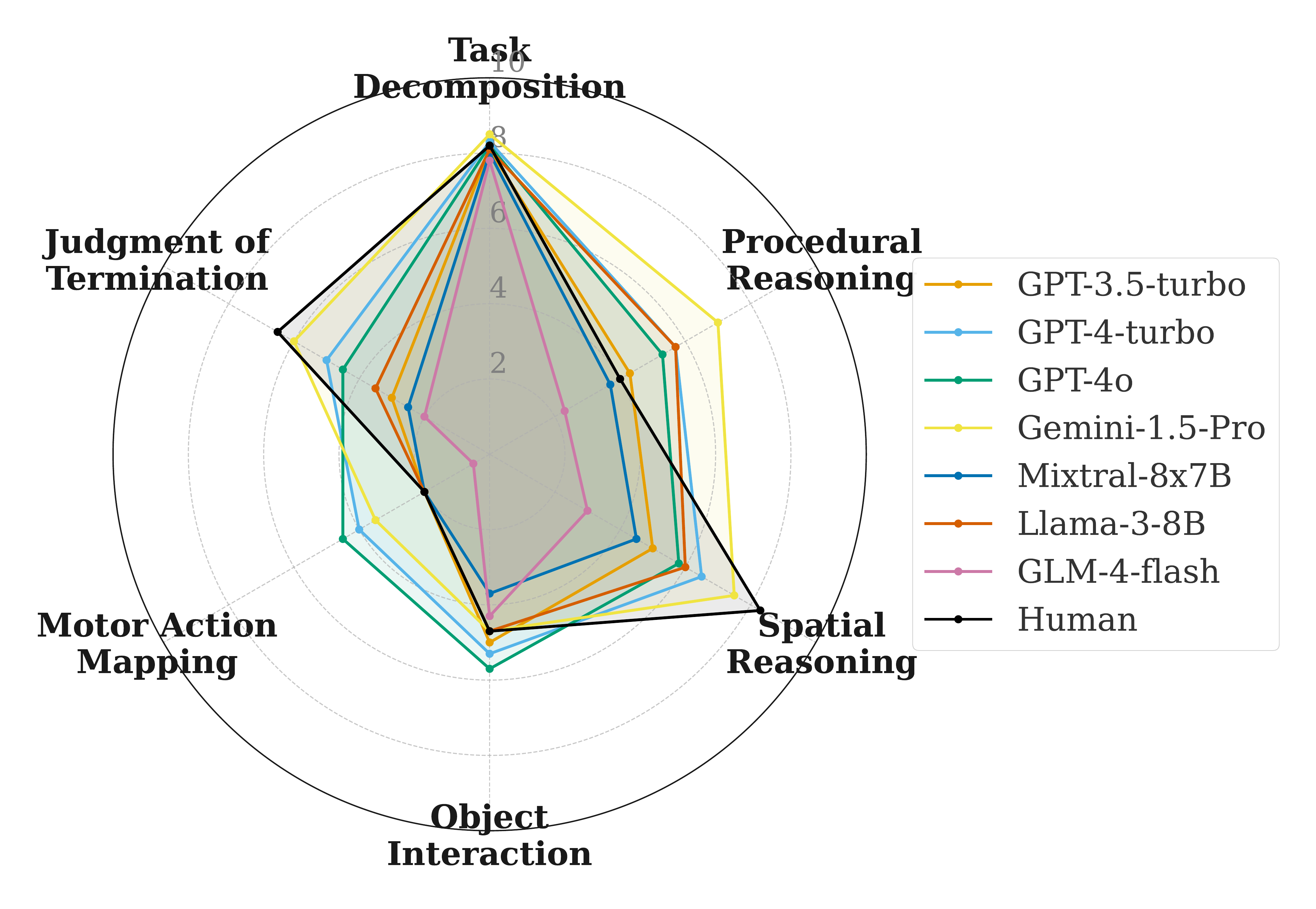}
\caption{Cognitive capabilities of LLMs and humans in translating semantic actions to VR device manipulations. Higher scores (0-10 scale) indicate stronger abilities.}
\label{fig:capabilities_radar}
\end{wrapfigure}
To contextualize our findings, we compare LLM performance against human baselines across our evaluation metrics. As shown in Tables~\ref{tab:overall_performance_1} and~\ref{tab:overall_performance_2}, state-of-the-art LLMs demonstrate competitive performance with humans on several key dimensions.
Our results reveal a nuanced performance landscape. While top-performing models (Gemini-1.5-Pro, GPT-4o) achieve comparable or superior NSAS scores relative to humans in certain games (e.g., Vivecraft: 0.931-0.938 vs. 0.935 for humans), human participants maintain a decisive advantage in Sequential Order Preservation, particularly for games requiring complex interaction sequences. In Half-Life: Alyx, humans achieve only 0.090 SOP compared to the best model performance of 0.209 (Gemini-1.5-Pro), yet this reflects the challenging nature of the task rather than superior model performance, both humans and models struggle with the intricate procedural requirements of this game.

Analysis of performance variance across games (Table~\ref{tab:cross_game_variation}) reveals striking similarities between human and high-performing model behavior. The standard deviation of human performance (0.084) closely aligns with that of GPT-4 (0.074) and Mixtral-8x7B (0.070), suggesting that both humans and advanced LLMs exhibit similar sensitivity patterns to game-specific interaction complexities. This convergence is particularly evident in structured environments like Vivecraft, where the consistency gap between humans and LLMs has substantially narrowed.
Figure~\ref{fig:capabilities_radar} illustrates the capability-wise performance comparison, revealing critical gaps in embodied reasoning. Humans maintain superior performance in spatial reasoning (8.3 vs. 7.5 for Gemini-1.5-Pro) and judgment of termination conditions (6.5 vs. 6.0). These differences are statistically significant ($p < 0.05$, Wilcoxon signed-rank test) and persist across all evaluated models. This performance gap suggests that while LLMs have achieved remarkable progress in understanding VR interaction semantics, they lack the grounded physical intuition that humans naturally apply when reasoning about three-dimensional manipulations and determining action completion states.

The convergence of human and LLM performance on certain metrics, coupled with persistent gaps in spatial and termination reasoning, indicates that current language models can effectively decompose VR tasks but struggle with aspects requiring embodied experience. This finding has important implications for the development of future VR-capable AI systems, suggesting the need for training paradigms that better incorporate spatial and physical reasoning capabilities.

%% file: Sections/6_Related_Work.tex
\section{Related Work}

Recent work has explored the use of large language models (LLMs) as generalist agents for embodied and interactive reasoning tasks. In robotics, \textit{SayCan}~\cite{ahn2022can} and \textit{PaLM-E}~\cite{driess2023palm} combine LLMs with affordance-based skill models or multimodal inputs to plan and execute robot actions in real-world settings. These methods demonstrate that LLMs can decompose high-level goals into actionable steps when grounded in sensory input and executable primitives. Similar capabilities have been investigated in virtual domains such as Minecraft through agents like \textit{Voyager}~\cite{wang2023voyager} and simulation platforms like \textit{MineDojo}~\cite{fan2022minedojo}, which showcase in-context learning and autonomous skill acquisition by prompting LLMs to generate and refine code or sub-goals based on environmental feedback. However, these systems are typically tuned for code-level or symbolic outputs and do not focus on physical device manipulation or spatially grounded motor control as required in VR environments.

Task decomposition and procedural reasoning have been studied extensively via prompting strategies such as Chain-of-Thought~\cite{wei2022chain} and ReAct~\cite{yao2022react}, which interleave reasoning with action selection to improve coherence in multi-step planning. LLMs have also been used to generate structured action sequences or API calls from natural instructions in domains like household tasks~\cite{shridhar2020alfred} and scientific procedures~\cite{wang2022scienceworld}. Code-as-policy paradigms~\cite{liang2022code} show that LLMs can output executable policy code that integrates logical control flow, enabling conditional and iterative actions. However, these approaches often abstract away the complexity of physical or spatial execution, making them less suited for evaluating embodied skills involving real-time device input, object affordances, or 3D spatial constraints.

To assess grounded reasoning, several benchmarks have been proposed in interactive settings. Animal-AI~\cite{mecattaf2024little} evaluates embodied cognition through physics-based tasks adapted from animal intelligence experiments, highlighting LLMs' partial competence in navigation, tool use, and physical causality. Similarly, platforms like ALFWorld~\cite{shridhar2021alfworld} and ScienceWorld~\cite{wang2022scienceworld} test instruction-following via text or symbolic interfaces, while MacGyver-style tasks~\cite{tian2024macgyver} probe object-use innovation in constrained settings. These works underscore known limitations in spatial reasoning, persistence, and tool-use generalization among LLMs. 

Concurrently, capability-oriented embodied evaluations for multimodal/embodied LLMs have emerged. EmbodiedBench~\cite{embodiedbench2025} unifies high- and low-level tasks with fine-grained error taxonomies, while VLABench~\cite{vlabench2024} targets long-horizon, language-conditioned manipulation for VLA policies; the Embodied Agent Interface (EAI)~\cite{eai2024} standardizes modules and metrics for step-level diagnostics. In parallel, GUI/OS/mobile control benchmarks, including OSWorld~\cite{osworld2024}, SPA-Bench~\cite{spabench2024}, WebArena~\cite{webarena2023}, Mind2Web~\cite{mind2web2023}, AndroidEnv~\cite{androidenv2021}, and AppAgent/AppAgent~v2~\cite{appagent2023,appagentv2_2024}, evaluate precise, procedure-level device interactions (click/tap/drag/typing) with programmatic success checks, offering a complementary view of grounded action competence. On the robotics side, VLA policies such as RT-1/RT-2~\cite{rt12022,rt22023} and OpenVLA~\cite{openvla2024} map visual observations and language to action tokens, improving semantic generalization in manipulation; large-scale 3D suites like Habitat~2.0/HAB~\cite{habitat2_2021}, BEHAVIOR-1K/OmniGibson~\cite{behavior1k2024}, and CALVIN~\cite{calvin2021} stress long-horizon rearrangement and physics. 

In contrast, our benchmark, \methodname, targets the translation of semantic goals into fine-grained, physically grounded VR device manipulations, enabling a more precise step-level analysis of embodied cognitive abilities critical for real-world interaction.

%% file: Sections/7_Conclusion.tex
\section{Conclusion}

We present \methodname, a comprehensive benchmark evaluating seven state-of-the-art LLMs on their ability to translate semantic actions into VR device manipulations across 262 scenarios from four popular VR games. Our evaluation reveals that while advanced models like Gemini-1.5-Pro demonstrate strong task decomposition capabilities (NSAS > 0.8), they exhibit significant weaknesses in motor action mapping and procedural reasoning, with Sequential Order Preservation scores often below 0.3 even in the best cases. Few-shot examples dramatically improve procedural understanding (10-20x increase in SOP scores) but provide limited benefit for exact step matching, suggesting that in-context learning helps models understand action relationships but cannot fully bridge the gap in physical manipulation reasoning. The pronounced performance variations across games and cognitive capabilities indicate that current text-trained LLMs lack the embodied understanding necessary for reliable VR interaction, pointing to the need for multimodal training approaches that incorporate spatial, visual, and haptic information. These findings highlight fundamental limitations in how language models represent physical interactions and suggest that achieving human-level VR manipulation capabilities will require architectural innovations beyond scaling current approaches, with important implications for the development of embodied AI systems in virtual and augmented reality applications.

%% file: Sections/Appendix.tex
\clearpage

\appendix

\section{Preliminaries on Virtual Reality}
\label{sec:background}

Virtual Reality (VR) represents a fundamentally distinct paradigm of human-computer interaction that transcends traditional interface boundaries. Unlike conventional computing systems that rely on indirect manipulation through keyboards, mice, and two-dimensional displays, VR creates immersive digital environments where users experience presence and embodiment. This paradigm shift necessitates a comprehensive understanding of both the technological infrastructure and the cognitive demands placed on users who must translate abstract intentions into concrete physical manipulations within virtual spaces.

The evolution of VR technology has progressed through several generations, from early tethered systems requiring substantial computational infrastructure to modern standalone devices that integrate processing, display, and tracking capabilities within compact form factors. Contemporary VR systems can be broadly categorized into three architectural approaches. PC-tethered headsets leverage external computational resources to deliver high-fidelity experiences with complex graphics and physics simulations. Standalone headsets, exemplified by devices like the Meta Quest series, incorporate integrated processors that balance performance with portability. Mobile-phone-based solutions represent an accessible entry point, utilizing smartphones as both display and processor, though with inherent limitations in tracking precision and computational capability.

The core hardware components enabling VR interaction form an integrated ecosystem of sensory input and output devices. Head-Mounted Displays (HMDs) serve as the primary visual interface, providing stereoscopic rendering that creates depth perception while simultaneously tracking head orientation and position through integrated sensors. This tracking enables natural viewing behaviors where users can examine virtual objects by physically moving their heads, mirroring real-world visual exploration patterns. Motion controllers, typically deployed in pairs to represent both hands, enable direct manipulation of virtual objects through a combination of positional tracking, button inputs, trigger mechanisms, and thumbstick controls. These devices must balance ergonomic considerations with functional complexity, providing sufficient input channels while maintaining intuitive operation. Spatial tracking systems, whether implemented through external sensors (outside-in tracking) or integrated cameras (inside-out tracking), monitor user movements with six degrees of freedom, capturing both translational and rotational motion to enable natural locomotion and interaction within virtual environments.

The ongoing evolution of VR hardware continues to introduce novel interaction modalities. Haptic gloves promise to deliver tactile feedback through actuators that simulate texture, resistance, and temperature. Full-body tracking systems capture skeletal motion to enable more nuanced avatar control and gesture recognition. Specialized peripherals, from steering wheels for racing simulations to weapon replicas for combat games, demonstrate the trend toward application-specific controllers that enhance immersion through physical affordances that match virtual interactions.

\subsection{Interaction Paradigms and Design Principles}
\label{subsec:interaction_paradigms}

The design of VR interaction paradigms represents a delicate balance between leveraging users' existing motor skills and introducing novel control schemes that exploit the unique capabilities of virtual environments. Direct manipulation forms the foundation of most VR interactions, where users employ hand controllers to simulate natural actions like grasping, throwing, and pushing. This approach capitalizes on users' lifetime of experience with physical object manipulation but requires careful calibration of virtual physics to match expectations. The mapping between controller inputs and virtual hand movements must account for the absence of tactile feedback, often employing visual or auditory cues to confirm successful interactions.

Ray-casting emerged as an elegant solution to the fundamental challenge of interacting with objects beyond physical reach. By projecting virtual rays from controllers, users can select, manipulate, and activate distant objects without locomotion. This technique exemplifies how VR interaction design often augments natural human capabilities rather than strictly simulating physical constraints. Advanced ray-casting implementations incorporate features like ray curvature for improved ergonomics, variable ray length based on context, and visual feedback mechanisms that indicate interaction possibilities.

Gesture recognition systems interpret temporal patterns of controller or hand movement as discrete commands, enabling a rich vocabulary of interactions without relying on button combinations. These systems must balance recognition accuracy with user comfort, avoiding gestures that cause fatigue or require precise movements difficult to perform consistently. Machine learning approaches have enhanced gesture recognition capabilities, allowing for more natural and varied input patterns while maintaining reliable detection rates.

Symbolic input mechanisms address scenarios where direct physical analogues are impractical or inefficient. Virtual keyboards present unique challenges in VR, as users lack tactile feedback and must rely on visual confirmation of key presses. Solutions range from laser-pointer selection of virtual keys to gesture-based text entry systems that map hand movements to characters. Voice commands offer an alternative input modality that bypasses manual interaction entirely, though they introduce considerations around recognition accuracy, latency, and social acceptability in shared spaces.

\subsection{Development Platforms and Technical Considerations}
\label{subsec:development}

The creation of VR applications relies on sophisticated development ecosystems that abstract hardware complexity while providing fine-grained control over interaction mechanics. Unity and Unreal Engine have emerged as dominant platforms, offering comprehensive toolsets that handle rendering pipelines, physics simulation, spatial audio, and cross-platform deployment. These engines provide specialized VR interaction frameworks that standardize common patterns like object grabbing, teleportation, and menu systems, significantly reducing development complexity.

Hardware software development kits (SDKs) serve as the bridge between high-level application logic and device-specific capabilities. Meta's OpenXR initiative represents an industry effort to standardize VR/AR interfaces, enabling applications to target multiple hardware platforms without extensive modifications. Platform-specific SDKs like SteamVR and Oculus SDK continue to play important roles, offering access to proprietary features and optimizations that enhance performance on particular hardware.

Technical constraints fundamentally shape VR interaction design decisions. Maintaining consistent frame rates above 72Hz (and preferably 90Hz or higher) prevents motion sickness and ensures responsive interactions. This performance requirement influences every aspect of application design, from polygon counts and texture resolution to the complexity of physics simulations. Tracking precision varies across hardware platforms and environmental conditions, necessitating interaction designs that accommodate occasional tracking losses or reduced accuracy. Developers must also consider the diverse computational capabilities across the VR ecosystem, implementing scalable solutions that provide acceptable experiences on entry-level hardware while leveraging the capabilities of high-end systems.

\subsection{Challenges in VR Interaction}
\label{subsec:challenges}

Despite remarkable technological progress, VR interaction continues to face fundamental challenges that impact user experience and limit application domains. The locomotion problem exemplifies the tension between physical and virtual spaces. While users may explore vast virtual environments, they remain constrained by finite physical play areas. Teleportation offers a practical solution but breaks immersion and can cause spatial disorientation. Artificial locomotion through thumbstick control risks motion sickness in susceptible users. More exotic solutions like omnidirectional treadmills or redirected walking techniques remain impractical for consumer applications.

The absence of comprehensive haptic feedback represents perhaps the most significant limitation in current VR systems. While controllers provide basic vibration feedback, they cannot simulate the rich tactile experiences of real-world interaction: the weight of objects, surface textures, temperature variations, or resistance to movement. This sensory gap creates a fundamental disconnect between visual expectations and physical sensations, requiring users to adapt their interaction strategies and often leading to reduced precision in manipulation tasks.

Interaction discoverability poses ongoing challenges as VR applications lack standardized interface conventions comparable to desktop or mobile platforms. Users encountering new VR experiences must often learn application-specific control schemes, gesture sets, and interaction patterns. The absence of persistent visual UI elements (to maintain immersion) exacerbates this challenge, as users cannot easily reference control schemes during gameplay. This lack of standardization increases cognitive load and creates barriers to entry for new users.

Precision manipulation tasks highlight the limitations of current tracking systems and input devices. Tasks requiring fine motor control, such as threading a virtual needle or manipulating small components, prove challenging due to tracking jitter, lack of physical surfaces for hand stabilization, and absence of tactile confirmation. These limitations restrict the types of applications suitable for VR and influence interaction design toward larger, more forgiving target sizes and simplified manipulation schemes.

\begin{figure}[!htbp]
  \centering
  \includegraphics[width=0.8\linewidth, height=0.20\textheight, keepaspectratio]{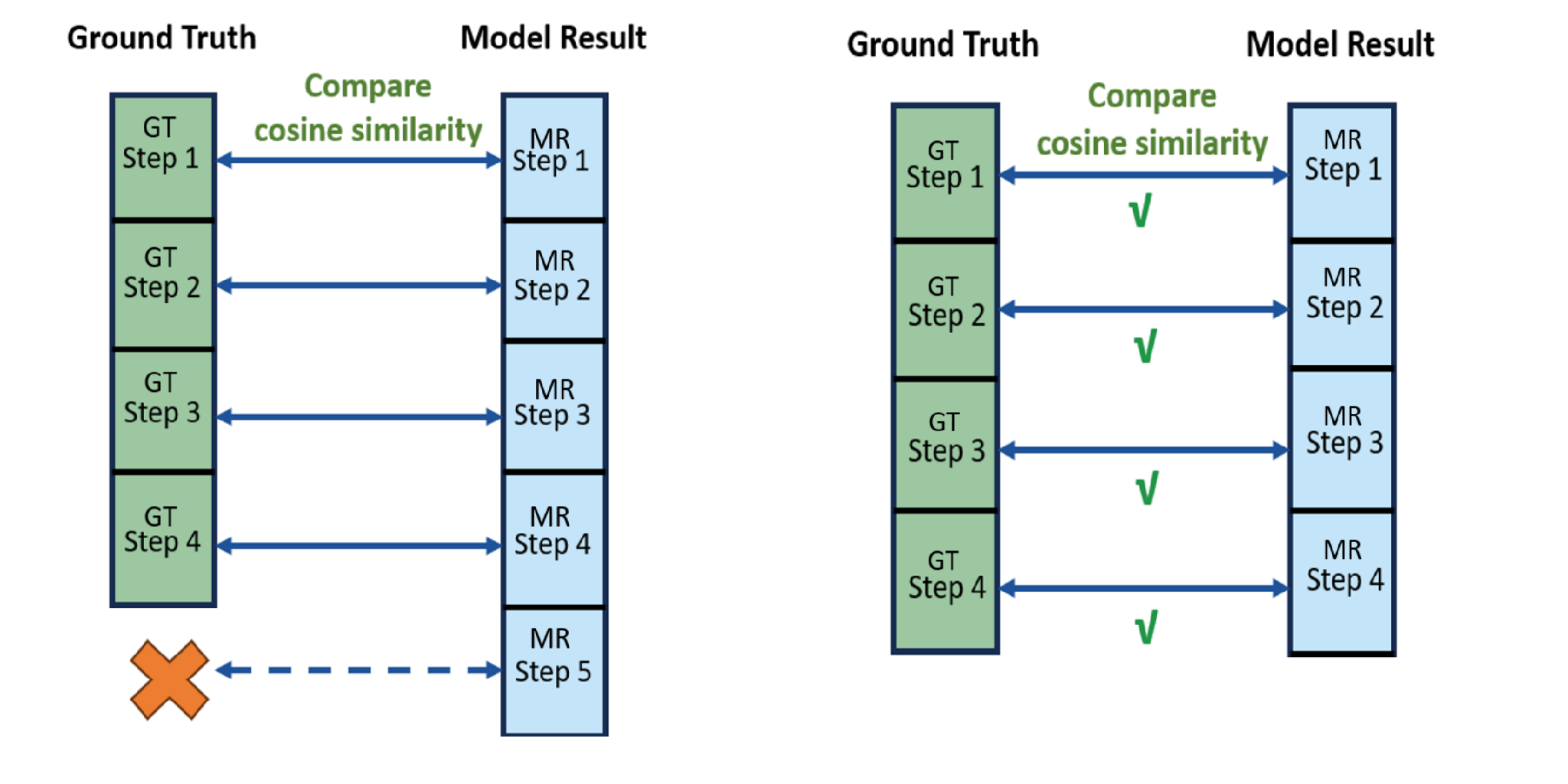}
  \caption{Overview of Strict Step-by-Step Matching (SSM) Calculation}
  \label{fig:ssm}
\end{figure}

\begin{figure}[!htbp]
  \vspace{0.8em}
  \centering
  \includegraphics[width=1.0\linewidth, height=0.48\textheight, keepaspectratio]{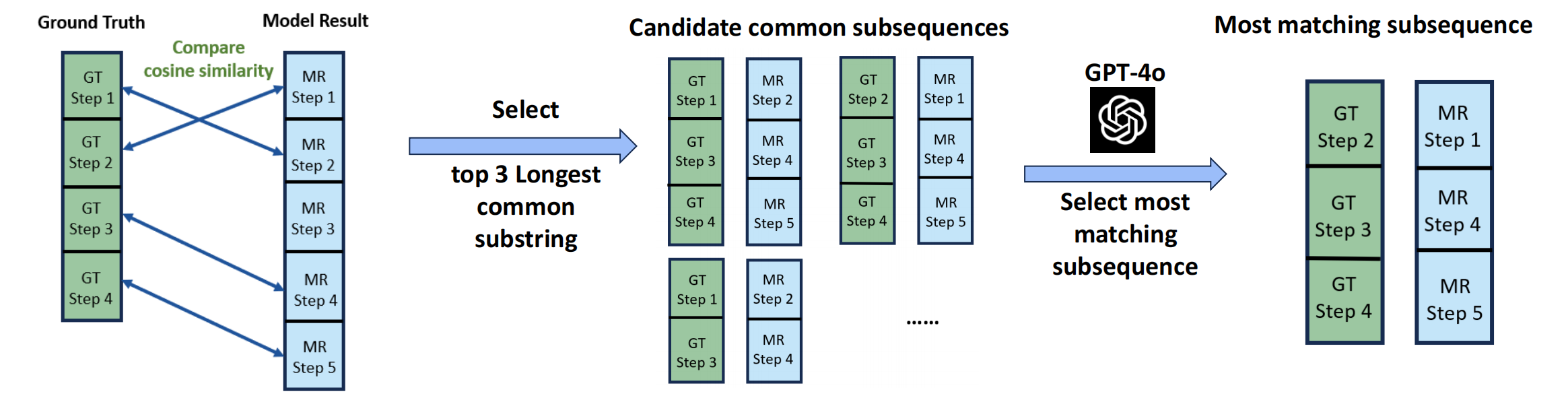}
  \caption{Overview of Common Subsequence Evaluation}
  \label{fig:cse}
\end{figure}

\section{Explanation of Evaluation Metrics}

\subsection{Strict Step-by-Step Matching (SSM)}

Figure~\ref{fig:ssm} illustrates the Strict Step-by-Step Matching (SSM) calculation process. SSM represents our most stringent evaluation metric, requiring exact correspondence between model-generated sequences and ground truth annotations. The calculation process operates as follows:

In the left panel, we observe a scenario where the ground truth contains 4 steps while the model result contains 5 steps. For SSM to register a match, two conditions must be satisfied: (1) the number of steps must be identical between ground truth and model output, and (2) each step must have a cosine similarity score above our threshold of 0.8387 with its corresponding ground truth step. In this example, the length mismatch alone disqualifies the sequence from being counted as correct, resulting in an SSM score of 0. The orange X symbol on the fifth model step visually indicates this length mismatch failure.

The right panel demonstrates a successful SSM match where both sequences contain 4 steps. Each model step is compared with its corresponding ground truth step using cosine similarity of their text embeddings. The green checkmarks indicate that all four step pairs exceed the similarity threshold, resulting in a successful match and contributing 1 to the SSM score. This metric's strictness explains why even high-performing models achieve relatively low SSM scores—any deviation in sequence length or individual step similarity results in complete failure for that sequence.

\subsection{Common Subsequence Evaluation}

Figure~\ref{fig:cse} details our Common Subsequence Evaluation approach, which underlies the Normalized Step Alignment Score (NSAS) and Sequential Order Preservation (SOP) metrics. This evaluation method provides more nuanced assessment than SSM by identifying partial matches and preserved ordering within sequences.

The process begins with comparing each step in the ground truth and model result sequences using cosine similarity, as shown by the crossing blue lines in the leftmost panel. Unlike SSM's strict position-based matching, this approach allows steps to match regardless of their positions in the sequences. The algorithm then identifies the top 3 longest common subsequences where matched steps maintain their relative ordering. 

In the example shown, multiple candidate subsequences are generated, each representing different ways steps from both sequences can be aligned while preserving order. The model (shown as GPT-4o) then selects the most matching subsequence based on the highest cumulative similarity scores. The final selected subsequence shows GT Steps 2, 3, and 4 matching with MR Steps 1, 4, and 5 respectively. This flexible matching approach allows the metrics to capture semantic correctness even when models include additional steps or present steps in slightly different positions.

The NSAS metric is calculated by considering the correctly matched steps (|C|), missing steps from ground truth (|M|), and additional steps in the model output (|A|), normalized by the total ground truth steps and scaled across the dataset. The SOP metric specifically evaluates whether matched steps maintain their sequential order, providing insight into the model's procedural reasoning capabilities.

\section{Detailed Experiment Results}
\label{sec:detailed_results}

This section provides comprehensive analysis of our experimental results, including detailed performance breakdowns across models, games, and experimental conditions. We present both aggregated metrics and fine-grained analyses that illuminate specific strengths and weaknesses in current LLMs' ability to reason about VR device manipulations.

\subsection{Overall Performance Analysis}

The table~\ref{tab:performance_llm_across_games} below presents a holistic view of model performance across all experimental conditions. The results reveal a clear performance hierarchy, with Gemini-1.5-Pro achieving the highest average Normalized Step Alignment Score (NSAS) of 0.845, followed closely by GPT-4o (0.832) and GPT-4 (0.824). Notably, even the best-performing models achieve relatively modest Strict Step-by-Step Matching (SSM) scores, with Gemini-1.5-Pro reaching only 8.7\% exact sequence matches. This discrepancy between NSAS and SSM scores indicates that while models can identify appropriate actions, they struggle with precise sequencing and complete reproduction of manipulation sequences.

The Sequential Order Preservation (SOP) scores reveal perhaps the most significant challenge facing current LLMs. Even top-performing models achieve SOP scores below 0.3, indicating difficulty in maintaining correct procedural ordering of steps. This limitation is particularly pronounced in zero-shot settings, where SOP scores approach zero for most models, suggesting that procedural reasoning for VR interactions requires exposure to examples rather than emerging from general language understanding.

Human performance provides an important baseline for contextualizing model achievements. While humans achieve comparable NSAS scores (0.817) to top LLMs, they show notably lower SOP scores (0.124) than leading models. This counterintuitive result reflects the challenging nature of the tasks even for experienced VR users and suggests that perfect procedural recall may be less important than adaptive problem-solving in real-world VR interaction.

\begin{table}[htbp!]
\caption{Performance of LLMs across VR Games (Best Few-Shot Setting)}
\label{tab:performance_llm_across_games}
\centering
\small
\begin{tabular}{lccccc}
\toprule
\textbf{Model} & \textbf{NSAS} & \textbf{SOP} & \textbf{SSC} & \textbf{SSM} & \textbf{Best FS} \\
\midrule
Gemini-1.5-Pro & 0.845 & 0.251 & 0.151 & 0.087 & 5 \\
GPT-4o & 0.832 & 0.291 & 0.190 & 0.135 & 5 \\
GPT-4 & 0.824 & 0.218 & 0.177 & 0.095 & 5 \\
LLaMA-3-8B & 0.823 & 0.283 & 0.200 & 0.040 & 5 \\
Human & 0.817 & 0.124 & 0.174 & 0.021 & - \\
Mixtral-8x7B & 0.790 & 0.123 & 0.142 & 0.039 & 5 \\
GPT-3.5 & 0.778 & 0.169 & 0.137 & 0.037 & 5 \\
GLM-4-Flash & 0.749 & 0.096 & 0.165 & 0.000 & 5 \\
\bottomrule
\end{tabular}
\end{table}
\subsection{Game-Specific Performance Patterns}

The table~\ref{tab:per_game} below reveals substantial variations in model performance across different VR games, highlighting how game design and interaction complexity influence LLM reasoning capabilities. Vivecraft consistently yields the highest performance across all models, with NSAS scores ranging from 0.909 to 0.938. This strong performance likely reflects the game's discrete, block-based interaction paradigm inherited from Minecraft, which provides clear action-object mappings that align well with linguistic descriptions.

In contrast, Into the Radius proves most challenging, with NSAS scores dropping to 0.618-0.698 across models. This game's emphasis on realistic physics simulation, complex inventory management, and weapon manipulation requires understanding of nuanced spatial relationships and multi-step procedures that current LLMs struggle to capture. The high standard deviation in performance (0.135 for GLM-4-flash) indicates inconsistent model behavior when confronting complex interaction scenarios.

Half-Life: Alyx and Moss: Book II occupy intermediate positions in the difficulty spectrum. Half-Life: Alyx's physics-based puzzles and combat scenarios require precise timing and spatial reasoning, reflected in extremely low SOP scores (0.022 for GPT-4o). Moss: Book II's third-person perspective and puzzle-platforming elements introduce unique challenges in translating camera-relative directions into controller movements, though models show more consistent performance than in Half-Life: Alyx.

\begin{table}[htbp!]
\caption{Performance comparison across different VR games (5-shot setting). We report NSAS scores (primary metric) and SOP scores (in parentheses).}
\label{tab:per_game}
\centering
\resizebox{0.75\columnwidth}{!}{
\begin{tabular}{l|cccc}
\toprule
\textbf{Model} & \textbf{Half-Life: Alyx} & \textbf{Radius} & \textbf{Moss} & \textbf{Vivecraft} \\
\midrule
GPT-3.5-turbo & 0.858 (0.123) & 0.662 (0.169) & 0.782 (0.169) & 0.922 (0.043) \\
GPT-4-turbo & 0.852 (0.125) & 0.693 (0.189) & 0.824 (0.218) & 0.927 (0.137) \\
GPT-4o & 0.804 (0.022) & 0.698 (0.291) & 0.824 (0.300) & 0.931 (0.190) \\
Gemini-1.5-Pro & 0.863 (0.209) & 0.682 (0.102) & 0.848 (0.265) & 0.938 (0.250) \\
Mixtral-8x7B & 0.839 (0.126) & 0.666 (0.123) & 0.756 (0.117) & 0.926 (0.060) \\
LLaMA-3-8B & 0.848 (0.126) & 0.644 (0.242) & 0.823 (0.283) & 0.929 (0.039) \\
GLM-4-flash & 0.836 (0.076) & 0.618 (0.096) & 0.749 (0.087) & 0.909 (0.000) \\
\midrule
Human & 0.845 (0.090) & 0.684 (0.148) & 0.817 (0.112) & 0.935 (0.122) \\
\bottomrule
\end{tabular}
}
\end{table}

\subsection{Impact of Few-Shot Learning}

The table~\ref{tab:fewshot_impact} below demonstrates the transformative effect of few-shot examples on model performance. The most dramatic improvements occur in SOP scores, which increase by factors of 10-20x from zero-shot to 5-shot settings. GPT-3.5-turbo exemplifies this pattern, improving from 0.036 to 0.226 in SOP F1 score, representing a 527.8\% relative gain. This massive improvement suggests that examples primarily help models understand the expected format and level of detail for procedural instructions rather than teaching fundamental VR interaction principles.

The diminishing returns pattern is consistent across models, with the largest gains occurring between zero-shot and 1-shot conditions. The jump from 3-shot to 5-shot provides minimal additional benefit, indicating that models quickly extract relevant patterns from limited examples. Gemini-1.5-Pro shows the most efficient few-shot learning, achieving top performance with fewer examples than competing models, suggesting superior in-context learning capabilities for procedural tasks.

Interestingly, few-shot examples have differential effects across game types. Complex games like Into the Radius show continued improvement with additional examples, while simpler environments like Vivecraft plateau quickly. This pattern indicates that few-shot learning is most beneficial when dealing with diverse interaction patterns and complex procedural sequences.

\begin{table}[t!]
\caption{Impact of few-shot examples on model performance. We report F1 scores for Sequential Order Preservation (SOP) across different number of examples. Higher is better.}
\label{tab:fewshot_impact}
\centering
\resizebox{0.73\columnwidth}{!}{
\begin{tabular}{l|cccc|c}
\toprule
\textbf{Model} & \textbf{Zero-shot} & \textbf{1-shot} & \textbf{3-shot} & \textbf{5-shot} & \textbf{Relative Gain (\%)} \\
\midrule
GPT-3.5-turbo & 0.036 & 0.096 & 0.190 & 0.226 & 527.8 \\
GPT-4-turbo & 0.102 & 0.171 & 0.254 & 0.301 & 195.1 \\
GPT-4o & 0.112 & 0.224 & 0.257 & 0.287 & 156.3 \\
Gemini-1.5-Pro & 0.085 & 0.187 & 0.260 & \textbf{0.330} & \textbf{288.2} \\
Mixtral-8x7B & 0.031 & 0.110 & 0.204 & 0.201 & 548.4 \\
LLaMA-3-8B & 0.090 & 0.165 & 0.254 & 0.299 & 232.2 \\
GLM-4-flash & 0.033 & 0.069 & 0.121 & 0.146 & 342.4 \\
\bottomrule
\end{tabular}
}
\end{table}

\subsection{Cognitive Capability Analysis}
The figure~\ref{fig:capabilities_radar} shows model performance across six cognitive dimensions, revealing distinct capability profiles. All models demonstrate strong task decomposition abilities (7.8-8.5), indicating that breaking down high-level goals into subtasks aligns well with LLMs' training on hierarchical text structures. Gemini-1.5-Pro leads in this dimension with a score of 8.5, though even smaller models like Mixtral-8x7B achieve respectable scores of 8.0.

Motor action mapping emerges as the most challenging capability across all models (0.5-4.5), highlighting the difficulty of translating abstract action concepts into specific button presses and controller movements. This limitation likely stems from the absence of embodied experience in text-based training data. GPT-4o performs best in this dimension but still falls far short of human-level capability, suggesting a fundamental gap in current architectures.

Procedural reasoning shows high variance across models (2.3-7.0), with Gemini-1.5-Pro again leading. The correlation between procedural reasoning scores and few-shot learning gains suggests that this capability can be partially addressed through examples, though the ceiling remains well below human performance. Spatial reasoning capabilities (4.8-7.5) reveal another significant gap, particularly evident in games requiring 3D navigation and object manipulation.

\subsection{Statistical Significance and Variance Analysis}

\begin{figure} [htbp!]
  \centering
\includegraphics[width=1.2\linewidth, height=0.9\textheight, keepaspectratio]{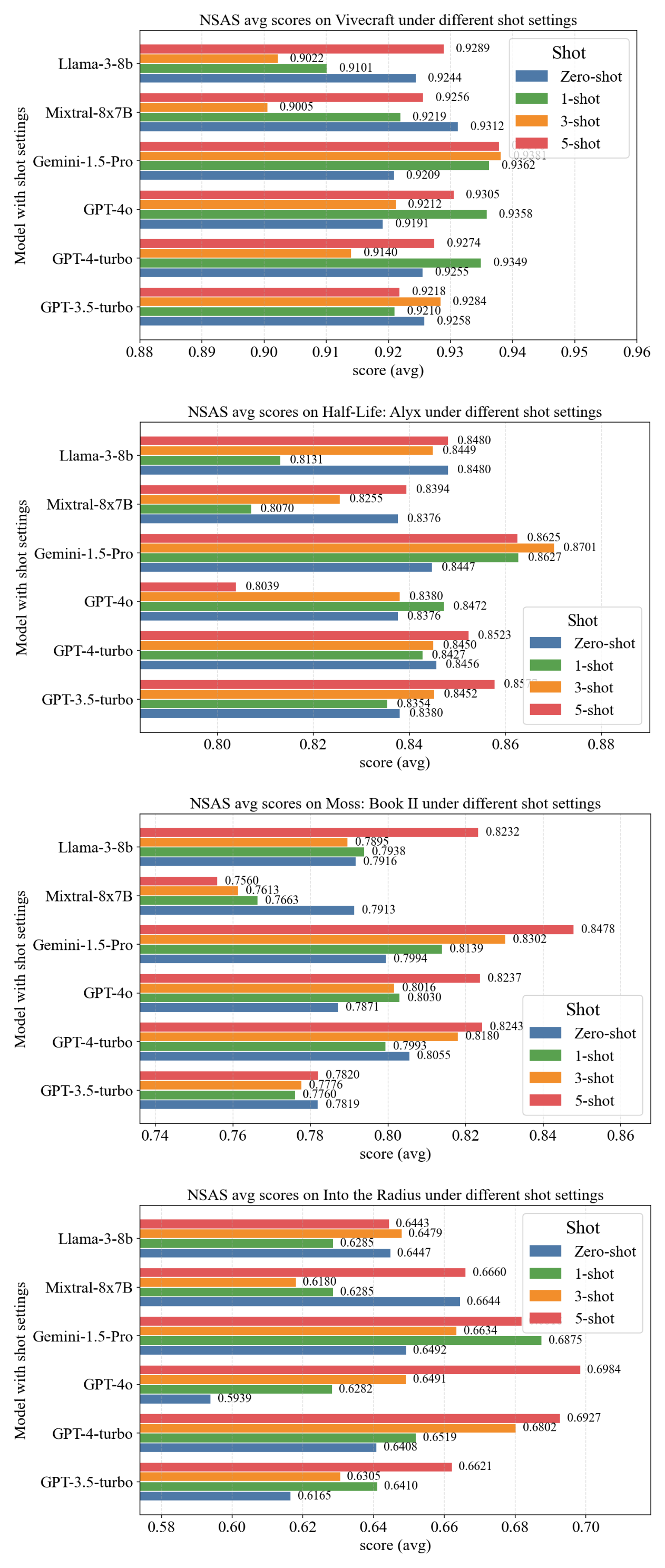}
  \caption{LLMs NSAS (avg) by Different Shot Setting Across Four VR Games}
  \vspace{0.8em}
  \label{fig:nsas_barh}
\end{figure}

\begin{figure} [htbp!]
\centering
  \includegraphics[width=1.2\linewidth, height=0.9\textheight, keepaspectratio]{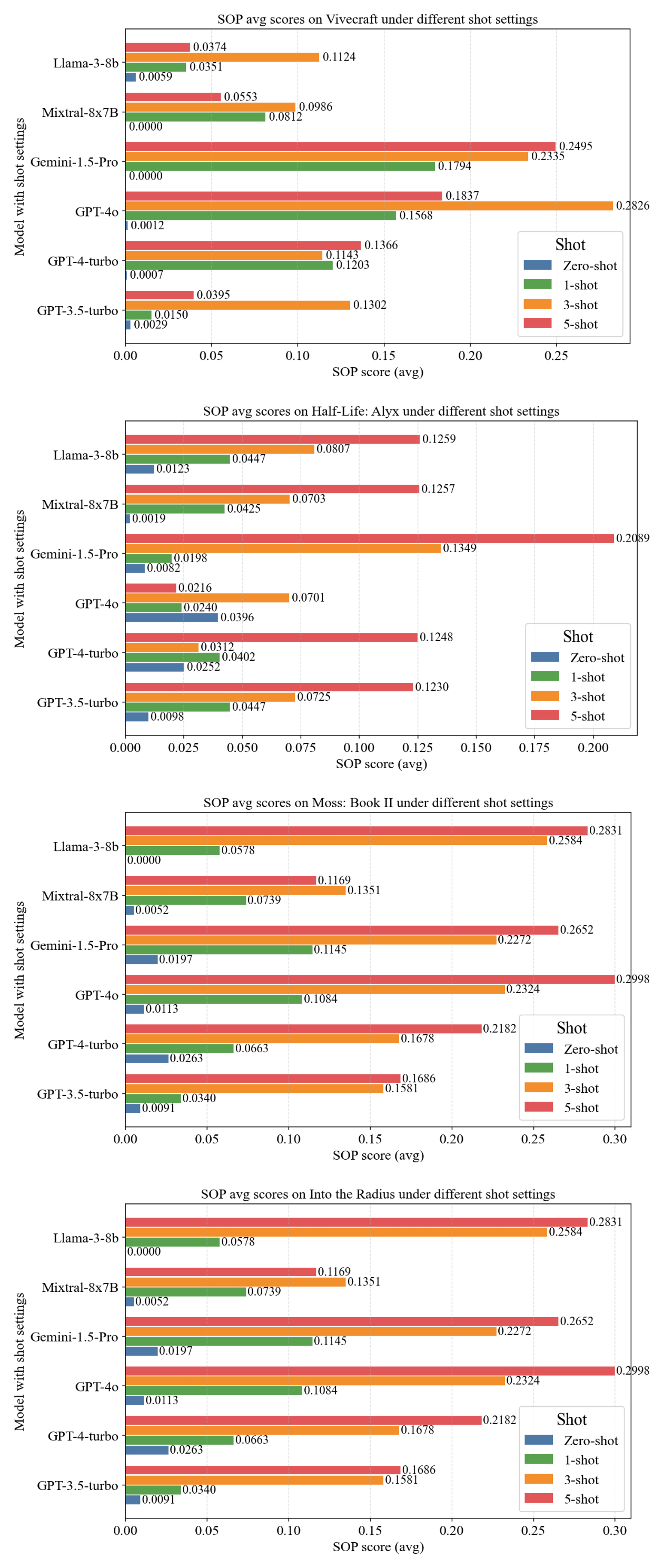}
  \caption{LLMs SOP (avg) by Different Shot Setting Across Four VR Games}
\vspace{0.8em}
\label{fig:sop_barh}
\end{figure}

\begin{figure} [htbp!]
\centering
  \includegraphics[width=1.2\linewidth, height=1.0\textheight, keepaspectratio]{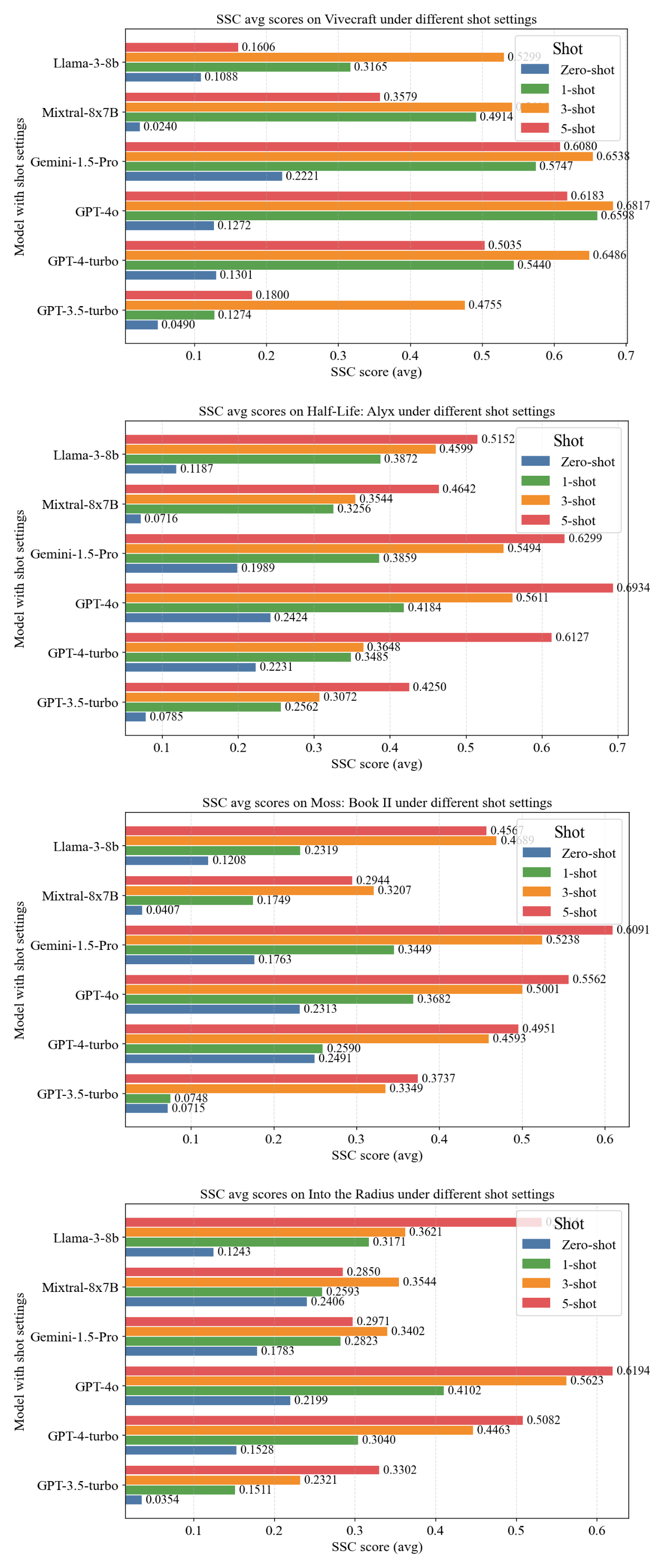}
  \caption{LLMs SSC (avg) by Different Shot Setting Across Four VR Games}
\vspace{0.8em}
\label{fig:ssc_barh}
\end{figure}

The tables~\ref{tab:vivecraft_nsas},~\ref{tab:vivecraft_sop},~\ref{tab:vivecraft_ssc},~\ref{tab:alyx_nsas},~\ref{tab:alyx_sop},~\ref{tab:alyx_ssc},~\ref{tab:moss_nsas},~\ref{tab:moss_sop},~\ref{tab:moss_ssc},~\ref{tab:radius_nsas},~\ref{tab:radius_sop}, and~\ref{tab:radius_ssc} below provide detailed statistical analyses of model performance, revealing important patterns in consistency and reliability. And the figures~\ref{fig:nsas_barh},~\ref{fig:sop_barh},~\ref{fig:ssc_barh} The standard deviation measurements across different games and shot settings illuminate which models maintain stable performance versus those exhibiting high variability. For instance, in Vivecraft, GPT-3.5-turbo shows remarkably consistent NSAS scores in zero-shot settings (std = 0.0248), but this consistency deteriorates with few-shot examples (std = 0.0734 at 3-shot), suggesting that additional examples introduce uncertainty in the model's approach to task completion.

The variance patterns differ significantly between metrics. NSAS scores generally show lower standard deviations (0.02-0.21 range) compared to SOP scores (0.00-0.34 range), indicating that models more consistently identify relevant steps than maintain proper ordering. This pattern is particularly pronounced in complex games like Into the Radius, where SOP standard deviations exceed 0.3 for several models in few-shot settings. Such high variance suggests that models employ different strategies across different runs, sometimes achieving correct ordering by chance rather than through systematic understanding.

Comparison with human variance provides crucial context for interpreting model stability. Human annotators show standard deviations comparable to mid-tier models (0.084 in cross-game performance), suggesting that some degree of variance is inherent to the task rather than a model limitation. However, humans maintain more consistent SOP performance (std = 0.029) compared to all models except Mixtral-8x7B, indicating more reliable procedural reasoning despite overall lower scores.

\begin{table}[h!]
    \caption{Average and standard deviation of Normalized Step Alignment Score (NSAS) scores comparison of LLMs on \textit{Vivecraft} under different shot settings.}
    \centering
    \resizebox{\textwidth}{!}{%
    \renewcommand{\arraystretch}{1.5}
    \begin{tabular}{l|cc|cc|cc|cc|cc|cc}
        \toprule
        \textbf{Model} & \multicolumn{2}{c|}{\textbf{GPT-3.5-turbo}} & \multicolumn{2}{c|}{\textbf{GPT-4-turbo}} & \multicolumn{2}{c|}{\textbf{GPT-4o}} & \multicolumn{2}{c|}{\textbf{Gemini-1.5-Pro}} & \multicolumn{2}{c|} {\textbf{Mixtral-8x7B}} &
        \multicolumn{2}{c}{\textbf{LLaMA-3-8b}} \\
        \midrule
        \textbf{Metrics} & \textbf{avg} & \textbf{std} & \textbf{avg} & \textbf{std} & \textbf{avg} & \textbf{std} & \textbf{avg} & \textbf{std} & \textbf{avg} & \textbf{std} & \textbf{avg} & \textbf{std} \\
        \midrule
        \textbf{Zero-shot}   & 0.9258 & 0.0248 & 0.9255 & 0.0238 & 0.9191 & 0.0306 & 0.9209 & 0.0334 & 0.9312 & 0.0207 & 0.9244 & 0.0329 \\
        \textbf{1-shot} & 0.921 & 0.0309 & 0.9349 & 0.0506 & 0.9358 & 0.0735 & 0.9362 & 0.0553 & 0.9219 & 0.0636 & 0.9101 & 0.0765 \\
        \textbf{3-shot} & 0.9284 & 0.0734 & 0.914 & 0.1167 & 0.9212 & 0.1115 & 0.9381 & 0.0781 & 0.9005 & 0.1125 & 0.9022 & 0.1051 \\
        \textbf{5-shot} & 0.9218 & 0.0385 & 0.9274 & 0.0674 & 0.9305 & 0.0689 & 0.9378 & 0.0708 & 0.9256 & 0.0477 & 0.9289 & 0.0364 \\
        \bottomrule
    \end{tabular}
    }
    
    \label{tab:vivecraft_nsas}
\end{table}

\begin{table}[h!]
    \caption{Average and standard deviation of Sequential Order Preservation (SOP) scores
comparison of LLMs on \textit{Vivecraft} under different shot settings.}
    \centering
    \resizebox{\textwidth}{!}{%
    \renewcommand{\arraystretch}{1.5}
    \begin{tabular}{l|cc|cc|cc|cc|cc|cc}
        \toprule
        \textbf{Model} & \multicolumn{2}{c|}{\textbf{GPT-3.5-turbo}} & \multicolumn{2}{c|}{\textbf{GPT-4-turbo}} & \multicolumn{2}{c|}{\textbf{GPT-4o}} & \multicolumn{2}{c|}{\textbf{Gemini-1.5-Pro}} & \multicolumn{2}{c|} {\textbf{Mixtral-8x7B}} &
        \multicolumn{2}{c}{\textbf{LLaMA-3-8b}} \\
        \midrule
        \textbf{Metrics} & \textbf{avg} & \textbf{std} & \textbf{avg} & \textbf{std} & \textbf{avg} & \textbf{std} & \textbf{avg} & \textbf{std} & \textbf{avg} & \textbf{std} & \textbf{avg} & \textbf{std} \\
        \midrule
        \textbf{Zero-Shot}  & 0.0029 & 0.0312 & 0.0007 & 0.0078 & 0.0012 & 0.0125 & 0.0 & 0.0 & 0.0 & 0.0 & 0.0059 & 0.0624 \\
        \textbf{1-shot}  & 0.015 & 0.0734 & 0.1203 & 0.2157 & 0.1568 & 0.2337 & 0.1794 & 0.291 & 0.0812 & 0.164 & 0.0351 & 0.1215 \\
        \textbf{3-shot}  & 0.1302 & 0.2352 & 0.1143 & 0.2136 & 0.2826 & 0.3417 & 0.2335 & 0.3417 & 0.0986 & 0.2024 & 0.1124 & 0.2026 \\
        \textbf{5-shot} & 0.0395 & 0.1226 & 0.1366 & 0.2388 & 0.1837 & 0.278 & 0.2495 & 0.3358 & 0.0553 & 0.158 & 0.0374 & 0.1371 \\
        \bottomrule
    \end{tabular}
    }

    \label{tab:vivecraft_sop}
\end{table}

\begin{table}[h!]
    \caption{Average and standard deviation of  Semantic Step Coverage (SSC) scores comparison of LLMs on \textit{Vivecraft} under different shot settings.}
    \centering
    \resizebox{\textwidth}{!}{%
    \renewcommand{\arraystretch}{1.5}
    \begin{tabular}{l|cc|cc|cc|cc|cc|cc}
        \toprule
        \textbf{Model} & \multicolumn{2}{c|}{\textbf{GPT-3.5-turbo}} & \multicolumn{2}{c|}{\textbf{GPT-4-turbo}} & \multicolumn{2}{c|}{\textbf{GPT-4o}} & \multicolumn{2}{c|}{\textbf{Gemini-1.5-Pro}} & \multicolumn{2}{c|} {\textbf{Mixtral-8x7B}} &
        \multicolumn{2}{c}{\textbf{LLaMA-3-8b}} \\
        \midrule
        \textbf{Metrics} & \textbf{avg} & \textbf{std} & \textbf{avg} & \textbf{std} & \textbf{avg} & \textbf{std} & \textbf{avg} & \textbf{std} & \textbf{avg} & \textbf{std} & \textbf{avg} & \textbf{std} \\
        \midrule
        \textbf{Zero-Shot} & 0.049 & 0.11 & 0.1301 & 0.1443 & 0.1272 & 0.1511 & 0.2221 & 0.2151 & 0.024 & 0.0999 & 0.1088 & 0.1549\\
        \textbf{1-shot}& 0.1274 & 0.1988 & 0.544 & 0.3672 & 0.6598 & 0.3322 & 0.5747 & 0.3359 & 0.4914 & 0.3617 & 0.3165 & 0.3415 \\
        \textbf{3-shot} & 0.4755 & 0.3526 & 0.6486 & 0.3373 & 0.6817 & 0.3204 & 0.6538 & 0.3373 & 0.5414 & 0.37 & 0.5299 & 0.3785 \\
        \textbf{5-shot} & 0.18 & 0.2337 & 0.5035 & 0.3772 & 0.6183 & 0.3416 & 0.608 & 0.3546 & 0.3579 & 0.3555 & 0.1606 & 0.2605 \\
        \bottomrule
    \end{tabular}
    }
    
    \label{tab:vivecraft_ssc}
\end{table}

\begin{table}[h!]
    \caption{Average and standard deviation of Normalized Step Alignment Score (NSAS) scores comparison of LLMs on \textit{Half-Life: Alyx} under different shot settings}
    \centering
    \resizebox{\textwidth}{!}{%
    \renewcommand{\arraystretch}{1.5}
    \begin{tabular}{l|cc|cc|cc|cc|cc|cc}
        \toprule
        \textbf{Model} & \multicolumn{2}{c|}{\textbf{GPT-3.5-turbo}} & \multicolumn{2}{c|}{\textbf{GPT-4-turbo}} & \multicolumn{2}{c|}{\textbf{GPT-4o}} & \multicolumn{2}{c|}{\textbf{Gemini-1.5-Pro}} & \multicolumn{2}{c|} {\textbf{Mixtral-8x7B}} &
        \multicolumn{2}{c}{\textbf{LLaMA-3-8b}} \\
        \midrule
        \textbf{Metrics} & \textbf{avg} & \textbf{std} & \textbf{avg} & \textbf{std} & \textbf{avg} & \textbf{std} & \textbf{avg} & \textbf{std} & \textbf{avg} & \textbf{std} & \textbf{avg} & \textbf{std} \\
        \midrule
        \textbf{Zero-Shot} & 0.838 & 0.0413 & 0.8456 & 0.0366 & 0.8376 & 0.0424 & 0.8447 & 0.032 & 0.8376 & 0.0331 & 0.848 & 0.0317 \\
        \textbf{1-shot} & 0.8354 & 0.0582 & 0.8427 & 0.0489 & 0.8472 & 0.0629 & 0.8627 & 0.0482 & 0.807 & 0.1099 & 0.8131 & 0.1289 \\
        \textbf{3-shot} & 0.8452 & 0.0551 & 0.845 & 0.0467 & 0.838 & 0.0757 & 0.8701 & 0.0603 & 0.8255 & 0.0819 & 0.8449 & 0.0707 \\
        \textbf{5-shot}  & 0.8577 & 0.0773 & 0.8523 & 0.0613 & 0.8039 & 0.0694 & 0.8625 & 0.0691 & 0.8394 & 0.0834 & 0.848 & 0.0976 \\
        \bottomrule
    \end{tabular}
    }
    
    \label{tab:alyx_nsas}
\end{table}

\begin{table}[h!]
    \caption{Average and standard deviation of Sequential Order Preservation (SOP) scores comparison of LLMs on \textit{Half Life: Alyx} under different shot settings.}
    \centering
    \resizebox{\textwidth}{!}{%
    \renewcommand{\arraystretch}{1.5}
    \begin{tabular}{l|cc|cc|cc|cc|cc|cc}
        \toprule
        \textbf{Model} & \multicolumn{2}{c|}{\textbf{GPT-3.5-turbo}} & \multicolumn{2}{c|}{\textbf{GPT-4-turbo}} & \multicolumn{2}{c|}{\textbf{GPT-4o}} & \multicolumn{2}{c|}{\textbf{Gemini-1.5-Pro}} & \multicolumn{2}{c|} {\textbf{Mixtral-8x7B}} &
        \multicolumn{2}{c}{\textbf{LLaMA-3-8b}} \\
        \midrule
        \textbf{Metrics} & \textbf{avg} & \textbf{std} & \textbf{avg} & \textbf{std} & \textbf{avg} & \textbf{std} & \textbf{avg} & \textbf{std} & \textbf{avg} & \textbf{std} & \textbf{avg} & \textbf{std} \\
        \midrule
        \textbf{Zero-Shot} & 0.0098 & 0.0802 & 0.0252 & 0.1265 & 0.0396 & 0.1745 & 0.0082 & 0.0669 & 0.0019 & 0.0158 & 0.0123 & 0.0704 \\
        \textbf{1-shot} & 0.0447 & 0.0764 & 0.0402 & 0.1224 & 0.024 & 0.0733 & 0.0198 & 0.1263 & 0.0425 & 0.0816 & 0.0447 & 0.0967 \\
        \textbf{3-shot} & 0.0725 & 0.1159 & 0.0312 & 0.0725 & 0.0701 & 0.1261 & 0.1349 & 0.2187 & 0.0703 & 0.1094 & 0.087 & 0.1687 \\
        \textbf{5-shot} & 0.123 & 0.1834 & 0.1248 & 0.2382 & 0.0216 & 0.0809 & 0.2089 & 0.2938 & 0.1257 & 0.2409 & 0.1259 & 0.2385 \\
        \bottomrule
    \end{tabular}
    }

    \label{tab:alyx_sop}
\end{table}

\begin{table} [htbp!]
    \caption{Average and standard deviation of  Semantic Step Coverage (SSC) scores comparison of LLMs on \textit{Half Life: Alyx} under different shot settings.}
    \centering
    \resizebox{\textwidth}{!}{%
    \renewcommand{\arraystretch}{1.5}
    \begin{tabular}{l|cc|cc|cc|cc|cc|cc}
        \toprule
        \textbf{Model} & \multicolumn{2}{c|}{\textbf{GPT-3.5-turbo}} & \multicolumn{2}{c|}{\textbf{GPT-4-turbo}} & \multicolumn{2}{c|}{\textbf{GPT-4o}} & \multicolumn{2}{c|}{\textbf{Gemini-1.5-Pro}} & \multicolumn{2}{c|} {\textbf{Mixtral-8x7B}} &
        \multicolumn{2}{c}{\textbf{LLaMA-3-8b}} \\
        \midrule
        \textbf{Metrics} & \textbf{avg} & \textbf{std} & \textbf{avg} & \textbf{std} & \textbf{avg} & \textbf{std} & \textbf{avg} & \textbf{std} & \textbf{avg} & \textbf{std} & \textbf{avg} & \textbf{std} \\
        \midrule
        \textbf{Zero-Shot} & 0.0785 & 0.1843 & 0.2231 & 0.2089 & 0.2424 & 0.2111 & 0.1989 & 0.1982 & 0.0716 & 0.1187 & 0.1662 & 0.172\\
        \textbf{1-shot}& 0.2562 & 0.235 & 0.3485 & 0.2336 & 0.4184 & 0.2413 & 0.3859 & 0.2654 & 0.3256 & 0.1934 & 0.3872 & 0.2058\\
        \textbf{3-shot} & 0.3072 & 0.2444 & 0.3648 & 0.2414 & 0.5611 & 0.229 & 0.5494 & 0.2887 & 0.3544 & 0.2202 & 0.4599 & 0.2371 \\
        \textbf{5-shot} & 0.425 & 0.2814 & 0.6127 & 0.2856 & 0.6934 & 0.2359 & 0.6299 & 0.315 & 0.4642 & 0.2957 & 0.5152 & 0.2708 \\
        \bottomrule
    \end{tabular}
    }
    \label{tab:alyx_ssc}
\end{table}

\begin{table} [htbp!]
    \caption{Normalized Step Alignment Score (NSAS) scores comparison of LLMs on \textit{ Moss: Book II} under different shot settings}
    \centering
    \resizebox{\textwidth}{!}{%
    \renewcommand{\arraystretch}{1.5}
    \begin{tabular}{l|cc|cc|cc|cc|cc|cc}
        \toprule
        \textbf{Model} & \multicolumn{2}{c|}{\textbf{GPT-3.5-turbo}} & \multicolumn{2}{c|}{\textbf{GPT-4-turbo}} & \multicolumn{2}{c|}{\textbf{GPT-4o}} & \multicolumn{2}{c|}{\textbf{Gemini-1.5-Pro}} & \multicolumn{2}{c|} {\textbf{Mixtral-8x7B}} &
        \multicolumn{2}{c}{\textbf{LLaMA-3-8b}} \\
        \midrule
        \textbf{Metrics} & \textbf{avg} & \textbf{std} & \textbf{avg} & \textbf{std} & \textbf{avg} & \textbf{std} & \textbf{avg} & \textbf{std} & \textbf{avg} & \textbf{std} & \textbf{avg} & \textbf{std} \\
        \midrule
        \textbf{Zero-Shot}  & 0.7819 & 0.0403 & 0.8055 & 0.0717 & 0.7871 & 0.0596 & 0.7994 & 0.0548 & 0.7913 & 0.0595 & 0.7916 & 0.0572 \\
        \textbf{1-shot}& 0.776 & 0.0616 & 0.7993 & 0.0771 & 0.803 & 0.0924 & 0.8139 & 0.0778 & 0.7663 & 0.0793 & 0.7938 & 0.0763 \\
        \textbf{3-shot} & 0.7776 & 0.0889 & 0.818 & 0.0925 & 0.8016 & 0.1242 & 0.8302 & 0.0935 & 0.7613 & 0.1341 & 0.7895 & 0.1371\\
        \textbf{5-shot} & 0.782 & 0.0952 & 0.8243 & 0.102 & 0.8237 & 0.1092 & 0.8478 & 0.1017 & 0.756 & 0.1469 & 0.8232 & 0.105\\
        \bottomrule
    \end{tabular}
    }
    \label{tab:moss_nsas}
\end{table}

\newpage
\begin{table} [htbp!]
    \caption{Average and standard deviation of Sequential Order Preservation (SOP) scores comparison of LLMs on \textit{Moss: Book II} under different shot settings.}
    \centering
    \resizebox{\textwidth}{!}{%
    \renewcommand{\arraystretch}{1.5}
    \begin{tabular}{l|cc|cc|cc|cc|cc|cc}
        \toprule
        \textbf{Model} & \multicolumn{2}{c|}{\textbf{GPT-3.5-turbo}} & \multicolumn{2}{c|}{\textbf{GPT-4-turbo}} & \multicolumn{2}{c|}{\textbf{GPT-4o}} & \multicolumn{2}{c|}{\textbf{Gemini-1.5-Pro}} & \multicolumn{2}{c|} {\textbf{Mixtral-8x7B}} &
        \multicolumn{2}{c}{\textbf{LLaMA-3-8b}} \\
        \midrule
        \textbf{Metrics} & \textbf{avg} & \textbf{std} & \textbf{avg} & \textbf{std} & \textbf{avg} & \textbf{std} & \textbf{avg} & \textbf{std} & \textbf{avg} & \textbf{std} & \textbf{avg} & \textbf{std} \\
        \midrule
        \textbf{Zero-Shot} & 0.0091 & 0.0581 & 0.0263 & 0.1533 & 0.0113 & 0.0494 & 0.0197 & 0.1029 & 0.0052 & 0.033 & 0.0 & 0.0\\
        \textbf{1-shot}& 0.034 & 0.1568 & 0.0663 & 0.2044 & 0.1084 & 0.2486 & 0.1145 & 0.2495 & 0.0739 & 0.1721 & 0.0578 & 0.1486 \\
        \textbf{3-shot} & 0.1581 & 0.242 & 0.1678 & 0.2522 & 0.2324 & 0.3185 & 0.2272 & 0.3324 & 0.1351 & 0.252 & 0.2584 & 0.3089 \\
        \textbf{5-shot} & 0.1686 & 0.244 & 0.2182 & 0.2801 & 0.2998 & 0.3062 & 0.2652 & 0.3596 & 0.1169 & 0.247 & 0.2831 & 0.3097 \\
        \bottomrule
    \end{tabular}
    }
    \label{tab:moss_sop}
\end{table}

\begin{table} [htbp!]
    \caption{Average and standard deviation of  Semantic Step Coverage (SSC) scores comparison of LLMs on \textit{Moss: Book II} under different shot settings.}
    \centering
    \resizebox{\textwidth}{!}{%
    \renewcommand{\arraystretch}{1.5}
    \begin{tabular}{l|cc|cc|cc|cc|cc|cc}
        \toprule
        \textbf{Model} & \multicolumn{2}{c|}{\textbf{GPT-3.5-turbo}} & \multicolumn{2}{c|}{\textbf{GPT-4-turbo}} & \multicolumn{2}{c|}{\textbf{GPT-4o}} & \multicolumn{2}{c|}{\textbf{Gemini-1.5-Pro}} & \multicolumn{2}{c|} {\textbf{Mixtral-8x7B}} &
        \multicolumn{2}{c}{\textbf{LLaMA-3-8b}} \\
        \midrule
        \textbf{Metrics} & \textbf{avg} & \textbf{std} & \textbf{avg} & \textbf{std} & \textbf{avg} & \textbf{std} & \textbf{avg} & \textbf{std} & \textbf{avg} & \textbf{std} & \textbf{avg} & \textbf{std} \\
        \midrule
        \textbf{Zero-Shot} & 0.0715 & 0.1792 & 0.2491 & 0.2844 & 0.2313 & 0.2567 & 0.1763 & 0.221 & 0.0407 & 0.0991 & 0.1208 & 0.1779\\
        \textbf{1-shot}& 0.0748 & 0.1719 & 0.259 & 0.2771 & 0.3682 & 0.3018 & 0.3449 & 0.3396 & 0.1749 & 0.2393 & 0.2319 & 0.293 \\
        \textbf{3-shot} & 0.3349 & 0.3069 & 0.4593 & 0.3309 & 0.5001 & 0.3444 & 0.5238 & 0.3738 & 0.3207 & 0.3105 & 0.4689 & 0.3399 \\
        \textbf{5-shot} & 0.3737 & 0.3213 & 0.4951 & 0.3373 & 0.5562 & 0.3319 & 0.6091 & 0.3476 & 0.2974 & 0.3385 & 0.4567 & 0.3173 \\
        \bottomrule
    \end{tabular}
    }
    \label{tab:moss_ssc}
\end{table}

\begin{table} [htbp!]
    \caption{Average and standard deviation of Normalized Step Alignment Score (NSAS) scores comparison of LLMs on VR game \textit{Into the Radius} under different shot settings}
    \centering
    \resizebox{\textwidth}{!}{%
    \renewcommand{\arraystretch}{1.5}
    \begin{tabular}{l|cc|cc|cc|cc|cc|cc}
        \toprule
        \textbf{Model} & \multicolumn{2}{c|}{\textbf{GPT-3.5-turbo}} & \multicolumn{2}{c|}{\textbf{GPT-4-turbo}} & \multicolumn{2}{c|}{\textbf{GPT-4o}} & \multicolumn{2}{c|}{\textbf{Gemini-1.5-Pro}} & \multicolumn{2}{c|} {\textbf{Mixtral-8x7B}} &
        \multicolumn{2}{c}{\textbf{LLaMA-3-8b}} \\
        \midrule
        \textbf{Metrics} & \textbf{avg} & \textbf{std} & \textbf{avg} & \textbf{std} & \textbf{avg} & \textbf{std} & \textbf{avg} & \textbf{std} & \textbf{avg} & \textbf{std} & \textbf{avg} & \textbf{std} \\
        \midrule
        \textbf{Zero-Shot} & 0.6165 & 0.0755 & 0.6408 & 0.0955 & 0.5939 & 0.1306 & 0.6492 & 0.1018 & 0.6644 & 0.0718 & 0.6447 & 0.0882 \\
        \textbf{1-shot} & 0.641 & 0.1177 & 0.6519 & 0.1421 & 0.6282 & 0.1687 & 0.6875 & 0.1159 & 0.6285 & 0.1684 & 0.6285 & 0.1346 \\
        \textbf{3-shot} & 0.6305 & 0.128 & 0.6802 & 0.1645 & 0.6491 & 0.2057 & 0.6634 & 0.1638 & 0.618 & 0.1633 & 0.6479 & 0.1606 \\
        \textbf{5-shot}  & 0.6621 & 0.1291 & 0.6927 & 0.1721 & 0.6984 & 0.2136 & 0.6818 & 0.1191 & 0.666 & 0.1495 & 0.6443 & 0.211\\
        \bottomrule
    \end{tabular}
    }

    \label{tab:radius_nsas}
\end{table}

\begin{table} [htbp!]
    \caption{Average and standard deviation of Sequential Order Preservation (SOP) scores comparison of LLMs on \textit{Into the Radius} under different shot settings.}
    \centering
    \resizebox{\textwidth}{!}{%
    \renewcommand{\arraystretch}{1.5}
    \begin{tabular}{l|cc|cc|cc|cc|cc|cc}
        \toprule
        \textbf{Model} & \multicolumn{2}{c|}{\textbf{GPT-3.5-turbo}} & \multicolumn{2}{c|}{\textbf{GPT-4-turbo}} & \multicolumn{2}{c|}{\textbf{GPT-4o}} & \multicolumn{2}{c|}{\textbf{Gemini-1.5-Pro}} & \multicolumn{2}{c|} {\textbf{Mixtral-8x7B}} &
        \multicolumn{2}{c}{\textbf{LLaMA-3-8b}} \\
        \midrule
        \textbf{Metrics} & \textbf{avg} & \textbf{std} & \textbf{avg} & \textbf{std} & \textbf{avg} & \textbf{std} & \textbf{avg} & \textbf{std} & \textbf{avg} & \textbf{std} & \textbf{avg} & \textbf{std} \\
        \midrule
        \textbf{Zero-Shot} & 0.0091 & 0.0581 & 0.0263 & 0.1533 & 0.0113 & 0.0494 & 0.0197 & 0.1029 & 0.0052 & 0.033 & 0.0 & 0.0\\
        \textbf{1-shot}& 0.034 & 0.1568 & 0.0663 & 0.2044 & 0.1084 & 0.2486 & 0.1145 & 0.2495 & 0.0739 & 0.1721 & 0.0578 & 0.1486 \\
        \textbf{3-shot} & 0.1581 & 0.242 & 0.1678 & 0.2522 & 0.2324 & 0.3185 & 0.2272 & 0.3324 & 0.1351 & 0.252 & 0.2584 & 0.3089 \\
        \textbf{5-shot} & 0.1686 & 0.244 & 0.2182 & 0.2801 & 0.2998 & 0.3062 & 0.2652 & 0.3596 & 0.1169 & 0.247 & 0.2831 & 0.3097 \\
        \bottomrule
    \end{tabular}
    }

    \label{tab:radius_sop}
\end{table}

\begin{table} [htbp!]
    \caption{Average and standard deviation of  Semantic Step Coverage (SSC) scores comparison of LLMs on \textit{Into the Radius} under different shot settings.}
    \centering
    \resizebox{\textwidth}{!}{%
    \renewcommand{\arraystretch}{1.5}
    \begin{tabular}{l|cc|cc|cc|cc|cc|cc}
        \toprule
        \textbf{Model} & \multicolumn{2}{c|}{\textbf{GPT-3.5-turbo}} & \multicolumn{2}{c|}{\textbf{GPT-4-turbo}} & \multicolumn{2}{c|}{\textbf{GPT-4o}} & \multicolumn{2}{c|}{\textbf{Gemini-1.5-Pro}} & \multicolumn{2}{c|} {\textbf{Mixtral-8x7B}} &
        \multicolumn{2}{c}{\textbf{LLaMA-3-8b}} \\
        \midrule
        \textbf{Metrics} & \textbf{avg} & \textbf{std} & \textbf{avg} & \textbf{std} & \textbf{avg} & \textbf{std} & \textbf{avg} & \textbf{std} & \textbf{avg} & \textbf{std} & \textbf{avg} & \textbf{std} \\
        \midrule
        \textbf{Zero-Shot} & 0.0354 & 0.0982 & 0.1528 & 0.1774 & 0.2199 & 0.1745 & 0.1783 & 0.2107 & 0.0406 & 0.0899 & 0.1243 & 0.1655\\
        \textbf{1-shot}& 0.1511 & 0.2246 & 0.304 & 0.2983 & 0.4102 & 0.2585 & 0.2823 & 0.3277 & 0.2593 & 0.3249 & 0.3171 & 0.269\\
        \textbf{3-shot}& 0.2321 & 0.2713 & 0.4463 & 0.3099 & 0.5623 & 0.2976 & 0.3402 & 0.3379 & 0.3544 & 0.2621 & 0.5319 & 0.2382\\
        \textbf{5-shot}& 0.3302 & 0.3115 & 0.5082 & 0.3063 & 0.6194 & 0.2877 & 0.2971 & 0.3187 & 0.285 & 0.3384 & 0.5314 & 0.2886\\
        \bottomrule
    \end{tabular}
    }
    \label{tab:radius_ssc}
\end{table}

\newpage

\FloatBarrier
\subsection{Cross-Game Generalization Patterns}

The cross-game performance analysis reveals important insights about model generalization capabilities. Models that perform well on one game do not necessarily maintain their advantage across others. For example, while GPT-4o achieves the highest SOP score in Into the Radius (0.291), it performs poorly in Half-Life: Alyx (0.022). This game-specific variation suggests that models may overfit to particular interaction patterns rather than developing general VR manipulation capabilities.

The "Game Gap" metric in the table~\ref{tab:cross_game_variation} quantifies this generalization challenge. Lower values indicate more consistent cross-game performance. Mixtral-8x7B achieves the lowest Game Gap (0.070), despite not leading in any individual game. This consistency might make it more suitable for applications requiring reliable performance across diverse VR experiences. In contrast, GPT-4o's high Game Gap (0.127) reflects its specialized strengths and weaknesses across different interaction paradigms.

Analysis of confusion patterns reveals that models struggle most when transitioning between games with different control schemes. The shift from Vivecraft's discrete block interactions to Half-Life: Alyx's continuous physics manipulation represents a fundamental change in how actions map to controller inputs. Models trained primarily on text lack the embodied experience to navigate these transitions smoothly, often applying inappropriate interaction patterns learned from one context to another.

\subsection{Temporal Dynamics in Sequential Tasks}

Detailed examination of step-by-step performance reveals how models handle temporal dependencies in VR interactions. Early steps in sequences generally show higher accuracy (NSAS > 0.9) across all models, with performance degrading for later steps. This degradation is particularly severe for steps that depend on the successful completion of previous actions. For instance, in a sequence like "pick up object, aim at target, throw object," models may correctly identify all three actions but fail to recognize that aiming requires successfully completing the pickup action first.

The SOP metric specifically captures these temporal dependencies, and the low scores across all models highlight a fundamental limitation in current architectures. Even with few-shot examples that demonstrate correct ordering, models struggle to internalize the causal relationships between steps. This suggests that improved performance may require architectural innovations that better capture temporal and causal reasoning, rather than simply scaling existing approaches.

Error analysis reveals common patterns in temporal mistakes. Models frequently suggest parallel actions that must be performed sequentially (e.g., "press trigger while reaching for object" when the trigger can only be meaningfully pressed after grasping). They also struggle with iterative processes, often omitting loop conditions or termination criteria. These patterns indicate that models lack an understanding of the physical constraints that govern VR interactions.

\subsection{Detailed Performance Tables and Visualizations}

The table~\ref{tab:overall_performance_2} provides granular data for researchers seeking to understand specific model behaviors. These tables reveal several noteworthy patterns. First, the relationship between different metrics is non-linear. High NSAS scores do not guarantee good SOP performance, and models with similar average scores may achieve them through different strengths. This multidimensional performance landscape suggests that selecting models for specific applications requires careful consideration of which capabilities are most critical.

The table~\ref{tab:overall_performance_2} illustrates the strict matching process, highlighting why SSM scores remain low even for generally capable models. The requirement for exact sequence length and step-by-step correspondence proves extremely demanding. Even minor variations in phrasing or step granularity result in match failures. This visualization helps explain why SSM may be overly strict for practical applications, where functional equivalence matters more than exact replication.

The table~\ref{tab:overall_performance_2} demonstrates the more nuanced evaluation approach that underlies our NSAS and SOP metrics. By identifying the longest common subsequences with semantic matching, these metrics better capture functional understanding while still penalizing significant deviations from ground truth. The visualization shows how models might achieve reasonable NSAS scores by identifying most relevant actions while still failing SOP evaluation due to ordering errors.

The heat maps of model performance across game-task combinations reveal clustering of difficulty. Certain task types (e.g., combat sequences in Half-Life: Alyx, inventory management in Into the Radius) consistently challenge all models, while others (e.g., block placement in Vivecraft) show near-ceiling performance. These patterns suggest that targeted improvements for specific interaction types might yield better results than general capability enhancement.

\subsection{Implications for Future Research}

The detailed experimental results paint a complex picture of current LLM capabilities and limitations in VR interaction reasoning. While models demonstrate competence in identifying relevant actions and decomposing high-level goals, they consistently struggle with the procedural and embodied aspects of VR interaction. The strong effect of few-shot examples suggests that current models possess latent capabilities that can be activated through appropriate prompting, but fundamental architectural limitations prevent them from achieving human-like understanding of physical manipulation sequences.

The high variance in performance across games and tasks indicates that robustness remains a significant challenge. Models that excel in one context may fail dramatically in another, limiting their practical applicability. This brittleness likely stems from the discrete nature of text-based training, which lacks the continuous, embodied experience that humans leverage when learning new physical tasks.

Moving forward, these results suggest several promising research directions. Multimodal models that incorporate visual and proprioceptive information alongside text may better capture the embodied nature of VR interactions. Explicit modeling of temporal and causal relationships could address the procedural reasoning gaps identified in our experiments. Finally, training on synthetic VR interaction data or through simulated embodiment might provide models with the experiential knowledge currently lacking in text-only approaches.

The detailed results also highlight the importance of comprehensive evaluation frameworks that assess multiple dimensions of capability. Single metrics fail to capture the complexity of VR interaction reasoning, and future benchmarks should continue to embrace multidimensional evaluation approaches that can identify specific strengths and weaknesses in model capabilities.

\section{Discussion, Limitations \& Broader Impacts}
\label{subsec:discussion}

Our investigation into LLMs' ability to translate semantic actions into VR device manipulations reveals both promising capabilities and fundamental limitations that reflect broader challenges in bridging linguistic understanding and embodied interaction. The relatively low Sequential Order Preservation (SOP) scores across all evaluated models indicate that current LLMs struggle with the temporal reasoning required for complex procedural tasks. This limitation suggests that while LLMs can identify relevant actions and understand their purposes, they lack the embodied experience necessary to accurately sequence physical manipulations.

The substantial performance variations across different VR games highlight how interaction complexity and consistency impact model performance. Games with standardized, discrete actions (like Vivecraft's block-based interactions) prove more amenable to LLM reasoning than those requiring nuanced controller movements or complex spatial reasoning (like Half-Life: Alyx). This pattern suggests that current language models may benefit from more structured representations of physical actions and explicit training on procedural sequences.

The significant improvement from few-shot examples demonstrates that LLMs possess latent capabilities for VR interaction reasoning that can be activated through appropriate prompting. However, the fact that performance plateaus with additional examples indicates fundamental architectural limitations rather than simple lack of exposure to relevant examples. This finding suggests that advances in VR-capable AI may require new training paradigms that incorporate spatial and temporal reasoning more directly.

From a broader perspective, this work carries important implications for the future of human-computer interaction and AI development. On the positive side, LLMs that can effectively reason about VR interactions could dramatically improve accessibility for users with motor impairments, enable more intuitive natural language interfaces for VR applications, and accelerate the development of intelligent tutoring systems for VR training scenarios. The potential transfer of these capabilities to robotic systems could enable more sophisticated human-robot collaboration in both virtual and physical environments.

However, we must also consider potential negative implications. As LLMs gain greater agency in controlling virtual (and potentially physical) systems, questions of safety, security, and user autonomy become paramount. The ability to translate high-level commands into detailed manipulation sequences could be exploited for unauthorized system control or social engineering attacks. Additionally, the computational resources required for training and deploying such models raise environmental concerns that must be balanced against their benefits.

The digital divide may be exacerbated as advanced VR-AI systems require substantial hardware investments and technical expertise. Ensuring equitable access to these technologies will require conscious effort from researchers, developers, and policymakers. Privacy concerns also emerge as these systems necessarily monitor and analyze detailed user movement patterns and interaction behaviors.

Moving forward, the field must pursue responsible development practices that prioritize user safety, privacy, and autonomy while advancing the technical capabilities of VR-AI systems. This includes developing robust evaluation frameworks that assess not only task performance but also failure modes, implementing transparent systems that users can understand and control, and ensuring that advances in VR interaction AI serve to augment rather than replace human agency in virtual environments.

\section{Large Language Models Usage Statement}

This work incorporated LLMs to aid in editorial refinement and linguistic improvement of the manuscript. The models provided assistance with stylistic enhancements and clarity optimization, including tasks such as rephrasing sentences and correcting grammatical errors.

We explicitly note that LLMs played no role in the conceptualization, theoretical development, or experimental design aspects of this research. The authors retain full responsibility for the entirety of the manuscript's content, including sections improved with LLM support. All LLM-assisted text has been carefully reviewed to ensure adherence to academic standards and ethical research practices.

%% file: iclr2026_conference.bbl
\begin{thebibliography}{46}
\providecommand{\natexlab}[1]{#1}
\providecommand{\url}[1]{\texttt{#1}}
\expandafter\ifx\csname urlstyle\endcsname\relax
  \providecommand{\doi}[1]{doi: #1}\else
  \providecommand{\doi}{doi: \begingroup \urlstyle{rm}\Url}\fi

\bibitem[web(2023)]{website:steam-app-store-vr}
{VR Content on Steam App Store}.
\newblock \url{https://store.steampowered.com/search/?vrsupport=401}, 2023.

\bibitem[Ahn et~al.(2022)Ahn, Brohan, Brown, Chebotar, Cortes, David, Finn, Fu, Gopalakrishnan, Hausman, et~al.]{ahn2022can}
Michael Ahn, Anthony Brohan, Noah Brown, Yevgen Chebotar, Omar Cortes, Byron David, Chelsea Finn, Chuyuan Fu, Keerthana Gopalakrishnan, Karol Hausman, et~al.
\newblock Do as i can, not as i say: Grounding language in robotic affordances.
\newblock \emph{arXiv preprint arXiv:2204.01691}, 2022.

\bibitem[Brohan et~al.(2022)Brohan, Brown, Carbajal, Chebotar, Dabis, Finn, Gopalakrishnan, Hausman, Herzog, Hsu, Ibarz, Ichter, Irpan, Jackson, Jesmonth, Joshi, Julian, Kalashnikov, Kuang, Leal, Lee, Levine, Lu, Malla, Manjunath, Mordatch, Nachum, Parada, Peralta, Perez, Pertsch, Quiambao, Rao, Ryoo, Salazar, Sanketi, Sayed, Singh, Sontakke, Stone, Tan, Tran, Vanhoucke, Vega, Vuong, Xia, Xiao, Xu, Xu, Yu, and Zitkovich]{rt12022}
Anthony Brohan, Noah Brown, Justice Carbajal, Yevgen Chebotar, Joseph Dabis, Chelsea Finn, Keerthana Gopalakrishnan, Karol Hausman, Alex Herzog, Jasmine Hsu, Julian Ibarz, Brian Ichter, Alex Irpan, Tomas Jackson, Sally Jesmonth, Nikhil Joshi, Ryan Julian, Dmitry Kalashnikov, Yuheng Kuang, Isabel Leal, Kuang-Huei Lee, Sergey Levine, Yao Lu, Utsav Malla, Deeksha Manjunath, Igor Mordatch, Ofir Nachum, Carolina Parada, Jodilyn Peralta, Emily Perez, Karl Pertsch, Jornell Quiambao, Kanishka Rao, Michael Ryoo, Grecia Salazar, Pannag Sanketi, Kevin Sayed, Jaspiar Singh, Sumedh Sontakke, Austin Stone, Clayton Tan, Huong Tran, Vincent Vanhoucke, Steve Vega, Quan Vuong, Fei Xia, Ted Xiao, Peng Xu, Sichun Xu, Tianhe Yu, and Brianna Zitkovich.
\newblock Rt-1: Robotics transformer for real-world control at scale, 2022.

\bibitem[Brohan et~al.(2023)Brohan, Brown, Carbajal, Chebotar, Chen, Choromanski, Ding, Driess, Dubey, Finn, Florence, et~al.]{rt22023}
Anthony Brohan, Noah Brown, Justice Carbajal, Yevgen Chebotar, Xi~Chen, Krzysztof Choromanski, Tianli Ding, Danny Driess, Avinava Dubey, Chelsea Finn, Pete Florence, et~al.
\newblock Rt-2: Vision-language-action models transfer web knowledge to robotic control, 2023.

\bibitem[CMGames(2019)]{intotheradius}
CMGames.
\newblock Into the {R}adius, 2019.
\newblock URL \url{https://www.into-the-radius.com/}.

\bibitem[Deng et~al.(2023)Deng, Gu, Zheng, Chen, Stevens, Wang, Sun, and Su]{mind2web2023}
Xiang Deng, Yu~Gu, Boyuan Zheng, Shijie Chen, Samuel Stevens, Boshi Wang, Huan Sun, and Yu~Su.
\newblock Mind2web: Towards a generalist agent for the web, 2023.

\bibitem[Driess et~al.(2023)Driess, Xia, Srinivas, Huang, M{\"u}ller, Mart{\'\i}n-Mart{\'\i}n, B{\"u}cheler, Du, Hausman, Tunyasuvunakool, et~al.]{driess2023palm}
Danny Driess, Fei Xia, Arjun Srinivas, Wenlong Huang, Julius M{\"u}ller, Roberto Mart{\'\i}n-Mart{\'\i}n, Tobias B{\"u}cheler, Yevgen~Chebotar Du, Karol Hausman, Saran Tunyasuvunakool, et~al.
\newblock Palm-e: An embodied multimodal language model.
\newblock \emph{arXiv preprint arXiv:2303.03378}, 2023.

\bibitem[Fan et~al.(2022)Fan, Wang, Jiang, Mandlekar, Yang, Zhu, Tang, Huang, Zhu, and Anandkumar]{fan2022minedojo}
Linxi Fan, Guanzhi Wang, Yunfan Jiang, Ajay Mandlekar, Yuncong Yang, Haoyi Zhu, Andrew Tang, De-An Huang, Yuke Zhu, and Anima Anandkumar.
\newblock Minedojo: Building open-ended embodied agents with internet-scale knowledge.
\newblock \emph{arXiv preprint arXiv:2206.08853}, 2022.

\bibitem[GLM et~al.(2024)GLM, Zeng, Xu, Wang, Zhang, Yin, Zhang, Rojas, Feng, Zhao, et~al.]{glm4}
Team GLM, Aohan Zeng, Bin Xu, Bowen Wang, Chenhui Zhang, Da~Yin, Dan Zhang, Diego Rojas, Guanyu Feng, Hanlin Zhao, et~al.
\newblock Chatglm: A family of large language models from glm-130b to glm-4 all tools.
\newblock \emph{arXiv preprint arXiv:2406.12793}, 2024.

\bibitem[Grattafiori et~al.(2024)Grattafiori, Dubey, Jauhri, Pandey, Kadian, Al-Dahle, Letman, Mathur, Schelten, Vaughan, et~al.]{llama3}
Aaron Grattafiori, Abhimanyu Dubey, Abhinav Jauhri, Abhinav Pandey, Abhishek Kadian, Ahmad Al-Dahle, Aiesha Letman, Akhil Mathur, Alan Schelten, Alex Vaughan, et~al.
\newblock The llama 3 herd of models.
\newblock \emph{arXiv preprint arXiv:2407.21783}, 2024.

\bibitem[Huang et~al.(2024)Huang, Wang, Li, Lam, Ren, Yuan, Jiao, Tu, and Lyu]{huang2024humanity}
Jen-tse Huang, Wenxuan Wang, Eric~John Li, Man~Ho Lam, Shujie Ren, Youliang Yuan, Wenxiang Jiao, Zhaopeng Tu, and Michael Lyu.
\newblock On the humanity of conversational ai: Evaluating the psychological portrayal of llms.
\newblock In \emph{The Twelfth International Conference on Learning Representations}, 2024.

\bibitem[Hurst et~al.(2024)Hurst, Lerer, Goucher, Perelman, Ramesh, Clark, Ostrow, Welihinda, Hayes, Radford, et~al.]{gpt4o}
Aaron Hurst, Adam Lerer, Adam~P Goucher, Adam Perelman, Aditya Ramesh, Aidan Clark, AJ~Ostrow, Akila Welihinda, Alan Hayes, Alec Radford, et~al.
\newblock Gpt-4o system card.
\newblock \emph{arXiv preprint arXiv:2410.21276}, 2024.

\bibitem[Jiang et~al.(2023)Jiang, Sablayrolles, Mensch, Bamford, Chaplot, de~las Casas, Bressand, Lengyel, Lample, Saulnier, et~al.]{mistral}
Albert~Q Jiang, Alexandre Sablayrolles, Arthur Mensch, Chris Bamford, Devendra~Singh Chaplot, Diego de~las Casas, Florian Bressand, Gianna Lengyel, Guillaume Lample, Lucile Saulnier, et~al.
\newblock Mistral 7b.
\newblock \emph{arXiv preprint arXiv:2310.06825}, 2023.

\bibitem[Kim et~al.(2024)Kim, Pertsch, Karamcheti, Xiao, Balakrishna, Nair, Rafailov, Foster, Lam, Sanketi, Vuong, Kollar, Burchfiel, Tedrake, Sadigh, Levine, Liang, and Finn]{openvla2024}
Moo~Jin Kim, Karl Pertsch, Siddharth Karamcheti, Ted Xiao, Ashwin Balakrishna, Suraj Nair, Rafael Rafailov, Ethan Foster, Grace Lam, Pannag Sanketi, Quan Vuong, Thomas Kollar, Benjamin Burchfiel, Russ Tedrake, Dorsa Sadigh, Sergey Levine, Percy Liang, and Chelsea Finn.
\newblock Openvla: An open-source vision-language-action model, 2024.

\bibitem[Lam et~al.(2025)Lam, Wang, Huang, and Lyu]{lam2025codecrash}
Man~Ho Lam, Chaozheng Wang, Jen-tse Huang, and Michael~R Lyu.
\newblock Codecrash: Stress testing llm reasoning under structural and semantic perturbations.
\newblock \emph{arXiv preprint arXiv:2504.14119}, 2025.

\bibitem[Lee et~al.(2024)Lee, Xia, Huang, Zhu, Zhang, and Lyu]{lee2024unified}
Cheryl Lee, Chunqiu~Steven Xia, Jen-tse Huang, Zhouruixin Zhu, Lingming Zhang, and Michael~R Lyu.
\newblock A unified debugging approach via llm-based multi-agent synergy.
\newblock \emph{arXiv preprint arXiv:2404.17153}, 2024.

\bibitem[Li et~al.(2024{\natexlab{a}})Li, Wong, Lingelbach, Mart\'{i}n-Mart\'{i}n, Fan, et~al.]{behavior1k2024}
Chengshu Li, Josiah Wong, Michael Lingelbach, Roberto Mart\'{i}n-Mart\'{i}n, Jim Fan, et~al.
\newblock Behavior-1k: A human-centered, embodied ai benchmark with 1{,}000 everyday activities and realistic simulation, 2024{\natexlab{a}}.

\bibitem[Li et~al.(2025)Li, Zhao, Wang, Wang, Zhou, Srivastava, Gokmen, Lee, Li, Zhang, Liu, Liang, Fei-Fei, Mao, and Wu]{eai2024}
Manling Li, Shiyu Zhao, Qineng Wang, Kangrui Wang, Yu~Zhou, Sanjana Srivastava, Cem Gokmen, Tony Lee, Li~Erran Li, Ruohan Zhang, Weiyu Liu, Percy Liang, Li~Fei-Fei, Jiayuan Mao, and Jiajun Wu.
\newblock Embodied agent interface: Benchmarking llms for embodied decision making, 2025.

\bibitem[Li et~al.(2024{\natexlab{b}})Li, Zhang, Yang, Fu, Cheng, Chen, Chen, and Wei]{appagentv2_2024}
Yanda Li, Chi Zhang, Wanqi Yang, Bin Fu, Pei Cheng, Xin Chen, Ling Chen, and Yunchao Wei.
\newblock Appagent v2: Advanced agent for flexible mobile interactions, 2024{\natexlab{b}}.

\bibitem[Liang et~al.(2022)Liang, Huang, Xia, Xu, Hausman, Ichter, Florence, and Zeng]{liang2022code}
Jacky Liang, Wenlong Huang, Fei Xia, Peng Xu, Karol Hausman, Brian Ichter, Pete Florence, and Andy Zeng.
\newblock Code as policies: Language model programs for embodied control.
\newblock \emph{arXiv preprint arXiv:2209.07753}, 2022.

\bibitem[Liang et~al.(2023)Liang, He, Jiao, Wang, Wang, Wang, Yang, Tu, and Shi]{liang2023encouraging}
Tian Liang, Zhiwei He, Wenxiang Jiao, Xing Wang, Yan Wang, Rui Wang, Yujiu Yang, Zhaopeng Tu, and Shuming Shi.
\newblock Encouraging divergent thinking in large language models through multi-agent debate.
\newblock \emph{arXiv preprint arXiv:2305.19118}, 2023.

\bibitem[Lu et~al.(2024)Lu, Bansal, Xia, Liu, Li, Hajishirzi, Cheng, Chang, Galley, and Gao]{lu2024mathvista}
Pan Lu, Hritik Bansal, Tony Xia, Jiacheng Liu, Chunyuan Li, Hannaneh Hajishirzi, Hao Cheng, Kai-Wei Chang, Michel Galley, and Jianfeng Gao.
\newblock Mathvista: Evaluating mathematical reasoning of foundation models in visual contexts.
\newblock In \emph{The Twelfth International Conference on Learning Representations}, 2024.

\bibitem[Mecattaf et~al.(2024)Mecattaf, Slater, Tešić, Prunty, Voudouris, and Cheke]{mecattaf2024little}
Matteo~G. Mecattaf, Ben Slater, Marko Tešić, Jonathan Prunty, Konstantinos Voudouris, and Lucy~G. Cheke.
\newblock A little less conversation, a little more action, please: Investigating the physical common-sense of llms in a 3d embodied environment.
\newblock \emph{arXiv preprint arXiv:2410.23242}, 2024.

\bibitem[Mees et~al.(2021)Mees, Hermann, Rosete-Beas, and Burgard]{calvin2021}
Oier Mees, Lukas Hermann, Erick Rosete-Beas, and Wolfram Burgard.
\newblock Calvin: A benchmark for language-conditioned policy learning for long-horizon robot manipulation tasks, 2021.

\bibitem[OpenAI(2022)]{gpt35}
OpenAI.
\newblock Introducing chatgpt.
\newblock \emph{OpenAI Blog Nov 30 2022}, 2022.
\newblock URL \url{https://openai.com/index/chatgpt/}.

\bibitem[OpenAI(2023)]{gpt4}
OpenAI.
\newblock Gpt-4 technical report.
\newblock \emph{arXiv preprint arXiv:2303.08774}, 2023.

\bibitem[Polyarc(2022)]{mossbookii}
Polyarc.
\newblock Moss: Book {II}, 2022.
\newblock URL \url{https://www.polyarcgames.com/games/moss-book-ii}.

\bibitem[Qin et~al.(2023)Qin, Zhang, Zhang, Chen, Yasunaga, and Yang]{qin2023chatgpt}
Chengwei Qin, Aston Zhang, Zhuosheng Zhang, Jiaao Chen, Michihiro Yasunaga, and Diyi Yang.
\newblock Is chatgpt a general-purpose natural language processing task solver?
\newblock \emph{arXiv preprint arXiv:2302.06476}, 2023.

\bibitem[Savva et~al.(2021)Savva, Kadian, Wijmans, Qian, Chang, et~al.]{habitat2_2021}
Manolis Savva, Abhishek Kadian, Erik Wijmans, Shengyi Qian, Angel Chang, et~al.
\newblock Habitat 2.0: Training home assistants to rearrange their habitat, 2021.

\bibitem[Shridhar et~al.(2020)Shridhar, Thomason, Gordon, Bisk, Han, Mottaghi, Zettlemoyer, and Fox]{shridhar2020alfred}
Mohit Shridhar, Jesse Thomason, Daniel Gordon, Yonatan Bisk, Winson Han, Roozbeh Mottaghi, Luke Zettlemoyer, and Dieter Fox.
\newblock Alfred: A benchmark for interpreting grounded instructions for everyday tasks.
\newblock In \emph{Proceedings of the IEEE/CVF Conference on Computer Vision and Pattern Recognition}, pp.\  10740--10749, 2020.

\bibitem[Shridhar et~al.(2021)Shridhar, Yuan, Côté, Bisk, Trischler, and Hausknecht]{shridhar2021alfworld}
Mohit Shridhar, Xingdi Yuan, Marc-Alexandre Côté, Yonatan Bisk, Adam Trischler, and Matthew Hausknecht.
\newblock Alfworld: Aligning text and embodied environments for interactive learning.
\newblock In \emph{Proceedings of the International Conference on Learning Representations (ICLR)}, 2021.
\newblock URL \url{https://arxiv.org/abs/2010.03768}.

\bibitem[Team et~al.(2024)Team, Georgiev, Lei, Burnell, Bai, Gulati, Tanzer, Vincent, Pan, Wang, et~al.]{gemini15}
Gemini Team, Petko Georgiev, Ving~Ian Lei, Ryan Burnell, Libin Bai, Anmol Gulati, Garrett Tanzer, Damien Vincent, Zhufeng Pan, Shibo Wang, et~al.
\newblock Gemini 1.5: Unlocking multimodal understanding across millions of tokens of context.
\newblock \emph{arXiv preprint arXiv:2403.05530}, 2024.

\bibitem[Tian et~al.(2024)Tian, Ravichander, Qin, Le~Bras, Marjieh, Peng, Choi, Griffiths, and Brahman]{tian2024macgyver}
Yufei Tian, Abhilasha Ravichander, Lianhui Qin, Ronan Le~Bras, Rami Marjieh, Nanyun Peng, Yejin Choi, Thomas~L Griffiths, and Faeze Brahman.
\newblock Macgyver: Are large language models creative problem solvers?
\newblock In \emph{Proceedings of the 2024 Conference of the North American Chapter of the Association for Computational Linguistics: Human Language Technologies}, pp.\  5303--5324, 2024.

\bibitem[Toyama et~al.(2021)Toyama, Hamel, Gergely, Comanici, Glaese, Ahmed, Jackson, Mourad, and Precup]{androidenv2021}
Daniel Toyama, Philippe Hamel, Anita Gergely, Gheorghe Comanici, Amelia Glaese, Zafarali Ahmed, Tyler Jackson, Shibl Mourad, and Doina Precup.
\newblock Androidenv: A reinforcement learning platform for android, 2021.

\bibitem[Valve(2020)]{halflifealyx}
Valve.
\newblock Half-{L}ife: Alyx, 2020.
\newblock URL \url{https://www.half-life.com/en/alyx/}.

\bibitem[Vivecraft(2013)]{vivecraft}
Vivecraft.
\newblock Vivecraft – {V}irtual {R}eality {M}inecraft for {S}team{VR}, 2013.
\newblock URL \url{https://www.vivecraft.org/}.

\bibitem[Wang et~al.(2023)Wang, Xie, Jiang, Mandlekar, Xiao, Zhu, Fan, and Anandkumar]{wang2023voyager}
Guanzhi Wang, Yuqi Xie, Yunfan Jiang, Ajay Mandlekar, Chaowei Xiao, Yuke Zhu, Linxi Fan, and Anima Anandkumar.
\newblock Voyager: An open-ended embodied agent with large language models.
\newblock \emph{arXiv preprint arXiv:2305.16291}, 2023.

\bibitem[Wang et~al.(2022)Wang, Khot, Sabharwal, and Clark]{wang2022scienceworld}
Yujia Wang, Tushar Khot, Ashish Sabharwal, and Peter Clark.
\newblock Scienceworld: Is your agent smarter than a 5th grader?
\newblock \emph{arXiv preprint arXiv:2203.07540}, 2022.

\bibitem[Wei et~al.(2022)Wei, Wang, Schuurmans, Bosma, Ichter, Xia, Chi, Le, and Zhou]{wei2022chain}
Jason Wei, Xuezhi Wang, Dale Schuurmans, Maarten Bosma, Brian Ichter, Fei Xia, Ed~Chi, Quoc~V Le, and Denny Zhou.
\newblock Chain-of-thought prompting elicits reasoning in large language models.
\newblock \emph{arXiv preprint arXiv:2201.11903}, 2022.

\bibitem[Xie et~al.(2024)Xie, Zhang, Chen, Li, Zhao, Cao, Hua, Cheng, Shin, Lei, Liu, Xu, Zhou, Savarese, Xiong, Zhong, and Yu]{osworld2024}
Tianbao Xie, Danyang Zhang, Jixuan Chen, Xiaochuan Li, Siheng Zhao, Ruisheng Cao, Toh~Jing Hua, Zhoujun Cheng, Dongchan Shin, Fangyu Lei, Yitao Liu, Yiheng Xu, Shuyan Zhou, Silvio Savarese, Caiming Xiong, Victor Zhong, and Tao Yu.
\newblock Osworld: Benchmarking multimodal agents for open-ended tasks in real computer environments, 2024.

\bibitem[Yang et~al.(2025)Yang, Chen, Zhang, Zhao, Qian, Wang, Wang, Koripella, Movahedi, Li, Ji, Zhang, and Zhang]{embodiedbench2025}
Rui Yang, Hanyang Chen, Junyu Zhang, Mark Zhao, Cheng Qian, Kangrui Wang, Qineng Wang, Teja~Venkat Koripella, Marziyeh Movahedi, Manling Li, Heng Ji, Huan Zhang, and Tong Zhang.
\newblock Embodiedbench: Comprehensive benchmarking multi-modal large language models for vision-driven embodied agents, 2025.

\bibitem[Yao et~al.(2022)Yao, Zhao, Yu, Du, Shafran, Narasimhan, and Cao]{yao2022react}
Shunyu Yao, Jeffrey Zhao, Dian Yu, Nan Du, Izhak Shafran, Karthik Narasimhan, and Yuan Cao.
\newblock React: Synergizing reasoning and acting in language models.
\newblock \emph{arXiv preprint arXiv:2210.03629}, 2022.

\bibitem[Zhang et~al.(2023)Zhang, Yang, Liu, Han, Chen, Huang, Fu, and Yu]{appagent2023}
Chi Zhang, Zhao Yang, Jiaxuan Liu, Yucheng Han, Xin Chen, Zebiao Huang, Bin Fu, and Gang Yu.
\newblock Appagent: Multimodal agents as smartphone users, 2023.

\bibitem[Zhang et~al.(2024{\natexlab{a}})Zhang, Xu, Liu, Yu, Li, Gao, Fei, Yin, Wu, Jiang, and Qiu]{vlabench2024}
Shiduo Zhang, Zhe Xu, Peiju Liu, Xiaopeng Yu, Yuan Li, Qinghui Gao, Zhaoye Fei, Zhangyue Yin, Zuxuan Wu, Yu-Gang Jiang, and Xipeng Qiu.
\newblock Vlabench: A large-scale benchmark for language-conditioned robotics manipulation with long-horizon reasoning tasks, 2024{\natexlab{a}}.

\bibitem[Zhang et~al.(2024{\natexlab{b}})Zhang, Qi, Ni, Yuan, Yang, et~al.]{spabench2024}
Zhaofeng Zhang, Yiyan Qi, Jinjie Ni, Jiayi Yuan, Fangkai Yang, et~al.
\newblock Spa-bench: A comprehensive benchmark for smartphone agent evaluation, 2024{\natexlab{b}}.

\bibitem[Zhou et~al.(2023)Zhou, Xu, Zhu, Zhou, Lo, Sridhar, Cheng, Ou, Bisk, Fried, Alon, and Neubig]{webarena2023}
Shuyan Zhou, Frank~F. Xu, Hao Zhu, Xuhui Zhou, Robert Lo, Abishek Sridhar, Xianyi Cheng, Tianyue Ou, Yonatan Bisk, Daniel Fried, Uri Alon, and Graham Neubig.
\newblock Webarena: A realistic web environment for building autonomous agents, 2023.

\end{thebibliography}
